\documentclass[pre,twocolumn,twoside,byrevtex,superscriptaddress,floatfix,nofootinbib]{revtex4-1}

\usepackage{currfile}
\usepackage[caption=false]{subfig}
\usepackage{afterpage}

\usepackage{graphicx,epsfig,verbatim,enumerate}
\usepackage{amssymb,amsmath}
\usepackage{ifthen}
\usepackage{float}
\usepackage{makecell}

\usepackage{bm}

\usepackage{placeins}

\usepackage{hyperref}

\usepackage{longtable}

\usepackage{mathtools}

\newboolean{twocolswitch}

\newcommand{\sindex}[1]{}
\newcommand{\nindex}[1]{}

\newcommand{\www}[1]{\url{#1}}

\usepackage{lettrine}

\usepackage{color}

\newcommand{\phiref}{\Phi^{(\text{ref})}}
\newcommand{\uptriangle}{\triangle}
\newcommand{\downtriangle}{\bigtriangledown}

\setboolean{twocolswitch}{true}

\begin{document}



\title{\protect
Generalized Word Shift Graphs:\\
A Method for Visualizing and Explaining Pairwise Comparisons Between Texts
}


\author{
    \firstname{Ryan}
    \surname{J. Gallagher}
}
\email{gallagher.r@northeastern.edu}
\affiliation{
    Network Science Institute, 
    Northeastern University, 
    Boston, MA 02115, USA
}
             
\author{
    \firstname{Morgan R.}
    \surname{Frank}
}
\affiliation{
    Department of Informatics and Networked Systems,
	University of Pittsburgh,
    Pittsburgh, PA 15260, USA
}
\affiliation{
    Connection Science,
    Massachusetts Institute of Technology,
    Cambridge, MA 02139, USA
}
\affiliation{
    Institute for Human-Centered Artifical Intelligence,
    Stanford University,
    Stanford, CA 94305, USA
}

\author{
    \firstname{Lewis}
    \surname{Mitchell}
}
\affiliation{
    School of Mathematical Sciences, 
    The University of Adelaide, 
    SA 5005, Australia}
    
\author{
    \firstname{Aaron J.}
    \surname{Schwartz}
}
\affiliation{
    Computational Story Lab,
    Vermont Complex Systems Center,
        \& Vermont Advanced Computing Core,
    The University of Vermont,
    Burlington, VT 05401, USA
}
\affiliation{
    Gund Institute for Environment 
        \& Rubenstein School of Environment and Natural Resources,
    The University of Vermont,
    Burlington, VT 05401, USA
}
\affiliation{
    Department of Ecology and Evolutionary Biology,
    University of Colorado at Boulder,
    Boulder, CO, 80309, USA
}

\author{
    \firstname{Andrew}
    \surname{J. Reagan}
}
\affiliation{
    MassMutual Data Science, Amherst, MA 01002, USA
}

\author{
    \firstname{Christopher}
    \surname{M. Danforth}
}
\affiliation{
    Computational Story Lab,
    Vermont Complex Systems Center,
        \& Vermont Advanced Computing Core,
    The University of Vermont,
    Burlington, VT 05401, USA
}
\affiliation{
    MassMutual Center of Excellence for Complex Systems and Data Science, 
        \& Department of Mathematics and Statistics,
    The University of Vermont,
    Burlington, VT 05401, USA
}
     
\author{
    \firstname{Peter}
    \surname{Sheridan Dodds}
}
\email{peter.dodds@uvm.edu}
\affiliation{
    Computational Story Lab,
    Vermont Complex Systems Center,
        \& Vermont Advanced Computing Core,
    The University of Vermont,
    Burlington, VT 05401, USA
}
\affiliation{
    MassMutual Center of Excellence for Complex Systems and Data Science, 
        \& Department of Mathematics and Statistics,
    The University of Vermont,
    Burlington, VT 05401, USA
}

\date{\today}

\begin{abstract}
  \protect
A common task in computational text analyses is to quantify how two corpora differ according to a measurement like word frequency, sentiment, or information content. However, collapsing the texts' rich stories into a single number is often conceptually perilous, and it is difficult to confidently interpret interesting or unexpected textual patterns without looming concerns about data artifacts or measurement validity. To better capture fine-grained differences between texts, we introduce generalized word shift graphs, visualizations which yield a meaningful and interpretable summary of how individual words contribute to the variation between two texts for any measure that can be formulated as a weighted average. We show that this framework naturally encompasses many of the most commonly used approaches for comparing texts, including relative frequencies, dictionary scores, and entropy-based measures like the Kullback-Leibler and Jensen-Shannon divergences. Through several case studies, we demonstrate how generalized word shift graphs can be flexibly applied across domains for diagnostic investigation, hypothesis generation, and substantive interpretation. By providing a detailed lens into textual shifts between corpora, generalized word shift graphs help computational social scientists, digital humanists, and other text analysis practitioners fashion more robust scientific narratives.
\end{abstract}

\maketitle

\section{Introduction}

News articles, audio transcripts, medical records, digitized archives, virtual libraries, computer logs, online memes, open-ended questionnaires, legislative proceedings, political manifestos, fan fiction, and poetry collections are just some of the many large-scale data sources that are readily available as text data \cite{lazer2009computational,salganik2019bit,grimmer2013text}. Computational methods help funnel what would be an otherwise overwhelming fire hose of raw text into coherent streams of social information \cite{grimmer2013text}. Social media text has allowed us to ask about the emotional pulse of large populations \cite{dodds2011temporal,mitchell2013geography}, the subtle adoption of community language by new members \cite{danescu2013no}, and the role of social bots in instigating political dialogue \cite{stella2018bots}. Digitized archives and collections have made it possible to observe the deliberative evolution of the French revolution \cite{barron2018individuals}, the lifespans of words across centuries \cite{petersen2012statistical,pechenick2017language}, and the social roles of characters in works of fiction \cite{sims2020measuring}. Song lyrics let us infer the latent emotions associated with musical chords \cite{kolchinsky2017minor}, legislative corpora show us the reuse of statutes across jurisdictions~\cite{funk2018spine}, and transcribed police body camera footage echos lived experiences of racial disparities in officer respect~\cite{voigt2017language}. Text as data fundamentally expands the number of social questions that we can ask across many different domains.

Computational methods for dealing with texts are abundant, but at the backbone of many of them is an intuitive concept: the weighted average. Weighted averages are a convenient tool because they are mathematically simple---it is easy to draw pairwise comparisons between texts by averaging over them in their entirety \cite{alajajian2017lexicocalorimeter,pechenick2015characterizing,gallagher2018divergent} or measure temporal trajectories by repeatedly averaging over time \cite{dodds2009c,barron2018individuals,reagan2016emotional,petersen2012statistical,pechenick2017language}. Domain knowledge \cite{nelson2018future,muddiman2019re} and social scientific constructs like sentiment \cite{dodds2011temporal,baylis2018weather}, morality \cite{brady2017emotion}, respect \cite{voigt2017language}, and hatefulness \cite{sood2012profanity,schmidt2017survey} can be integrated through the weights, making it easy to adapt average-based methods to new situations and focus them on particular questions of interest.

However, the simplicity of the weighted average is often one of its most significant drawbacks. Collapsing texts down to a single number introduces serious concerns about measurement validity because it is not always clear a mere weighted average can capture complex social phenomena \cite{loughran2011liability,grimmer2013text,nelson2018future,barbera2019automated}. Even if one accepts a particular weighted average as a conceptually valid measurement, that measure can still vary in unanticipated ways. The sheer abundance of language underlying computational text models \cite{zipf1949human,simon1955class} can cause a given measure to rise or fall due to the frequent appearance of a single set of key words \cite{pury2011automation} or an unexpected combination of frequent and less frequent words \cite{schmidt2012words,pechenick2015characterizing}. Further, the relative weights of those words may be highly context dependent \cite{loughran2011liability,grimmer2013text}---regional dialects \cite{munro2010crowdsourced,schwaiger2016assessing}, variations in slang \cite{bucholtz2007hella}, and other domain-specific usage \cite{loughran2011liability,hamilton2016inducing,baucom2013mirroring} all affect how appropriate it is to compare across weighted averages. Even if contextual weights can be derived for different sets of text, there are limited tools for comparing weighted averages beyond their aggregate value. While theory can provide guidance at times, it is a perilous path towards reliably interpreting text data if we do not have methods for interpreting the averages themselves.

We contend that these concerns can and should be addressed by systematically quantifying \emph{which} words contribute to the differences between two texts, and, importantly, \emph{how} they do so. To this end, we propose \emph{generalized word shift graphs}, horizontal bar charts which provide word-level explanations of how and why two texts differ across any measure derived from a weighted average. The framework that we propose generalizes previous formulations of word shifts \cite{dodds2011temporal,reagan2017sentiment} to account for how a word changes in both relative frequency and measurement, allowing us to unify a wide range of common measures under the same methodological banner, including dictionary scores, Shannon entropy, the Kullback-Leibler divergence, the Jensen-Shannon divergence, generalized entropies, and any other measure that can be written as a weighted average or difference in weighted averages.

Through a number of case studies, we show that generalized word shift graphs address many of the aforementioned issues: they unmask the internal workings of aggregate averages, enumerate exactly which words contribute to variation in a measure, account for context-dependent measurements across different settings, diagnose measurement issues during the research process, and provide an interpretable tool for validating, constructing, and presenting scientifically sound stories. We advocate for the use of generalized word shift graphs among computational social scientists, digital humanists, and other text analysis practitioners, and release open source code to encourage their uptake in the methodological toolkit for working with text as data.\footnote{We make our open source code for constructing generalized word shift graphs available at:\\ \url{https://github.com/ryanjgallagher/shifterator}.}

\section{Pairwise Comparisons Between Texts}

We first present a number of measures that are representative of the many different ways that two different texts can be quantitatively juxtaposed. As we show more explicitly later, all of these can be written as a weighted average or difference in weighted averages. For each measure, we provide guidance on the questions that it is most capable of answering, and the benefits and limitations of applying the measure to draw out differences between texts.

Here and onward we denote our two text corpora by $T^{(1)}$ and $T^{(2)}$. We consider the full vocabulary $\mathcal{T}$, composed of all the word types in either $\mathcal{T}^{(1)}$ or $\mathcal{T}^{(2)}$. Each word type $\tau$ in the vocabulary $\mathcal{T}$ appears with some frequency $f_\tau^{(i)}$ in each of the texts, where either $f_\tau^{(1)}$ or $f_\tau^{(2)}$ may be zero. We notate each type's normalized, relative frequency as $p_\tau^{(i)} = f_\tau^{(i)} / \sum_{\tau' \in \mathcal{T}} f_{\tau'}^{(i)}$. Unless otherwise specified, we use ``word'' to mean ``word type,'' where a ``word'' may be any $n$-gram or phrase as defined by the vocabulary, and not necessarily just a unigram.

\subsection{Relative Frequency}

One of the simplest and most common ways of identifying the most characteristic words of two texts is to compare how often each word appears in one text versus the other. That is, we can compute the difference in their relative frequencies,
\begin{equation}
    p_\tau^{(2)} - p_\tau^{(1)}.
    \nonumber
\end{equation}
As we can see, if the difference is positive then the word is relatively more common in $T^{(2)}$, if it is negative then it is more common in $T^{(1)}$, and if it is zero then it is equally common in both texts. We can rank words by the magnitude of this difference to produce a list of words that distinguish the texts from one another.

Comparing the relative frequency of words is adequate for a cursory pass of two texts, but it is less attuned to identifying subtle, but characteristic differences between them. Consider a word used frequently in both $T^{(1)}$ and $T^{(2)}$. Then the absolute difference $|p_\tau^{(2)} - p_\tau^{(1)}|$ has more potential for being large because $p_\tau^{(1)}$ and $p_\tau^{(2)}$ are themselves large. Yet, exactly because the word is frequently used, it is unlikely that the difference in usage will be surprising or substantively interesting. On the other hand, a less frequently used but more distinct word, can only have a difference as large as the maximum of $p_\tau^{(1)}$ and $p_\tau^{(2)}$, hindering its ability to rank highly. Comparing the relative frequencies of words puts more emphasis on differences between the most frequently used words, and less on the long, rich tail of word usage \cite{zipf1949human,simon1955class} that may leave more lexical clues to what characterizes the texts.   

\subsection{Entropy}

Shannon entropy accounts for both a word's relative frequency and its unexpectedness. If we let $P$ denote the entire normalized distribution of words in a text with vocabulary $\mathcal{T}$, then the (Shannon) entropy \cite{shannon1948mathematical} is given by
\begin{equation}
    H(P)
        =
            \sum_{\tau \in \mathcal{T}}
                p_\tau \log_2 \frac{1}{p_\tau}.
    \nonumber
\end{equation}
The entropy measures the unpredictability of a text: it is maximized if every word is equally likely to occur (i.e., $p_\tau = 1 / N$ for all $N$ words in the vocabulary), and minimized if only one word is used (i.e., $p_\tau = 1$ for a single word $\tau$ and 0 for all others). At the word level, the factor $\log_2 1 / p_\tau$ distinguishes a word's contribution to $H(P)$ from just its relative frequency $p_\tau$. This factor is known as a word's \emph{surprisal}---a word is more surprising if it is used relatively less. Another way of interpreting the entropy then is as the average surprisal of a text.

To compare two texts, we can consider the difference in their entropies,
\begin{equation}
    H\bigl(P^{(2)}\bigr) - H\bigl(P^{(1)}\bigr).
    \nonumber
\end{equation}
By considering the components of the sums, we can decompose the difference into the contribution from each word $\tau$,
\begin{equation}
\delta H_\tau
    = 
        p^{(2)}_\tau \log_2 \frac{1}{p^{(2)}_\tau}
                -
                p^{(1)}_\tau \log_2 \frac{1}{p^{(1)}_\tau}.
    \nonumber
\end{equation}
Like relative frequencies, we can order words by their absolute contribution to obtain a ranked list of the words that are most characteristic of each text. Unlike relative frequencies, each word's surprisal weights it inversely to its frequency. Generalized, or Tsallis, entropies \cite{havrda1967quantification} introduce a tunable parameter to further control how much consideration is given to rare and common words \cite{jost2006entropy,hill1973diversity,altmann2017generalized} (see Materials and Methods for details), and the Shannon entropy is a special limiting case that statistically balances between those that frequently and infrequently occur \cite{jost2006entropy,hill1973diversity}. Entropy has been particularly effective as an operationalization of diversity \cite{jost2006entropy}, where it has been used to measure textual diversities like the lexical diversity of online populations \cite{dodds2011temporal}, the hashtag diversity of online activism \cite{gallagher2018divergent}, and the information content diversity of search engine results \cite{steiner2020seek}.


\subsection{Kullback-Leibler Divergence}

At times we may want an asymmetric measure of how texts differ. For instance, we may want to measure how language evolved with respect to some reference point in the past \cite{barron2018individuals}, or compare the language of one person to that of an entire community \cite{danescu2013no}. For these cases, we distinguish between a \emph{reference} text and a \emph{comparison} text. If w let $P^{(1)}$ be the relative word frequency distribution of the reference text and $P^{(2)}$ be the distribution of the comparison, then the Kullback-Leibler divergence (KLD), or relative entropy, is defined as:
\begin{align}
    D^{(\text{KL})} 
        \bigl(
            &P^{(2)} || P^{(1)} 
        \bigr) \nonumber\\
            &=
                \sum_{\tau \in \mathcal{T}}
                    p_\tau^{(2)} \log_2 \frac{1}{p_\tau^{(1)}}
                    - p_\tau^{(2)} \log_2 \frac{1}{p_\tau^{(2)}}.
    \nonumber
\end{align}
The KLD is the average number of extra bits per word required to encode the words of text $T^{(2)}$ using an optimal coding scheme for $T^{(1)}$ instead of $T^{(2)}$.
As such, it shares a form similar to entropy where each word's contribution is the difference between the surprisal of the word in the reference and comparison, but, in contrast to entropy, both surprisals are weighted by the word's relative frequency in the comparison text. The KLD is a conceptually useful measure when we have a well-defined vocabulary and a meaningful reference distribution for comparison. However, if there is a single word that appears in the vocabulary of the comparison but not the reference (i.e., $p_\tau^{(2)} > 0$ and $p_\tau^{(1)} = 0$), then the KLD is infinite. This makes the KLD a brittle measure for comparing texts in general because it is only applicable if the comparison text uses a subset of words from the reference text's lexicon, which is very often not the case when comparing two distinct corpora. 

\subsection{Jensen-Shannon Divergence}

The Jensen-Shannon divergence (JSD) accounts for some of the shortcomings of the Kullback-Leibler divergence. The JSD compares the similarity of the word distributions by first constructing a probability distribution $M$ for some artificial hybrid text:
\begin{equation}
    M = \pi_1 P^{(1)} + \pi_2 P^{(2)}.
    \nonumber
\end{equation}
The mixture weights $\pi_1$ and $\pi_2$ must sum to 1 and are often set to be either equal, $\pi_1 = \pi_2 = 1/2$, or proportional to the number of word tokens in $T^{(1)}$ and $T^{(2)}$. The JSD is then computed as the average KLD from the mixture text,
\begin{align}
    D^{(\text{JS})} 
        \bigl( 
            &P^{(1)} || P^{(2)} 
        \big) \nonumber \\
            &=
                \pi_1 D^{(KL)}
                    \bigl(
                        P^{(1)} || M
                    \bigr)
                +
                \pi_2 D^{(KL)}
                    \bigl(
                        P^{(2)} || M
                    \bigr).
    \nonumber
\end{align}
By construction, the JSD is symmetric and does not infinitely diverge like the KLD because $M$ consists of the entire vocabulary of both texts. Conveniently, the JSD takes on a value of 0 if the texts are identical and a value of 1 if they have no words in common (as long as we are using base 2 logarithms). The individual contribution $\delta \text{JSD}_\tau$ of a word $\tau$ to the JSD is given by,
\begin{equation}
    \delta \text{JSD}_\tau
        =
            m_\tau \log \frac{1}{m_\tau}
            - \biggl( 
                \pi_1 p_\tau^{(1)} \log \frac{1}{p_\tau^{(1)}}
                + \pi_2 p_\tau^{(2)} \log \frac{1}{p_\tau^{(2)}}
            \biggr),
\label{eq:jsd-contribution}
\end{equation}
the (corpus-weighted) difference between the surprisal of the word in the average text and the average surprisal of the word in each observed text. Note, the contribution is always non-negative, and $\delta \text{JSD}_\tau = 0$ if and only if $p_\tau^{(1)} = p_\tau^{(2)}$. Like Shannon entropy, the JSD can be generalized to emphasize different regions of the word frequency distribution \cite{altmann2017generalized} (see Materials and Methods for details). The symmetric nature of the JSD has made it a useful tool for investigating cultural evolution across digitized collections \cite{pechenick2015characterizing}, charting fluctuations in the birth and death of words \cite{pechenick2017language}, and disentangling viewpoints in online political discussions \cite{gallagher2018divergent}

\subsection{Dictionary-Based Scores}

The measures that we have introduced so far all compare texts based on the relative frequencies of their words. The differences between them lie in how they weight each contribution, where those weights are themselves functions of word frequency. Very often though, we have external weights that we want to specify for each word. The most common example of this is dictionary-based sentiment analysis \cite{dodds2011temporal,mohammad2018obtaining,mohammad2018word}, where we have a dictionary of words and each word is assigned a weight or score according to its association with a particular emotion or feeling. Other dictionaries and lexicons have been curated to encode constructs like morality \cite{brady2017emotion}, respect \cite{voigt2017language}, profanity \cite{sood2012profanity}, and hatefulness \cite{schmidt2017survey}.

When we are equipped with dictionary scores, we can calculate the average score of each text as a whole and then compare them. If we have a single dictionary that prescribes a score $\phi_\tau$ for each word $\tau$ in the vocabulary $\mathcal{T}$, then the difference between the weighted averages $\Phi^{(1)}$ and $\Phi^{(2)}$ is
\begin{equation}
\delta \Phi
    =
        \sum_{\tau \in \mathcal{T}} \phi_\tau 
            \bigl( 
                p_\tau^{(2)} - p_\tau^{(1)} 
            \bigr).
    \nonumber
\end{equation}
When the dictionary does not cover the entire vocabulary (as is often the case), we typically subset the vocabulary to only words appearing in the dictionary. Like the other measures, we can use the linearity of the weighted averages to extract the contributions $\delta \Phi_\tau$ to the difference and rank them accordingly.

\section{Word Shift Graphs}
\label{sec:word-shift}

When using any weighted average for pairwise text comparison, we want to be able to interpret differences between measurements. Each of the measures that we have introduced can be decomposed into word-level contributions, and so we can identify \emph{which} words most account for the between-text variation. We would like to go further and explain \emph{how} each word contributes. Is one set of lyrics happier than another because it uses more positive words or because, instead, it uses less negative words? Does a social bot's language seem unpredictable because it uses a variety of surprising words or because it uses common words in a surprising way? To what extent do misogynistic internet communities not only use sexist slurs, but also associate other words with negative overtones? These are the kinds of qualitative and contextual questions that can be answered quantitatively through the word shift framework and visualized through word shift graphs.

\subsection{Word Shift Fundamentals}
\label{subsec.basicwordshifts}

We first revisit basic word shift graphs which we first introduced in ref.~\cite{dodds2009c} in the context of happiness measurements, and further developed in refs.~\cite{dodds2011temporal} and~\cite{dodds2015a}. Basic word shifts are for use when we have single set of scores unchanged across texts \cite{dodds2011temporal}, as is often the case for (but in no way limited to) standard dictionary-based sentiment analyses. We then generalize the word shift framework so that each text can be equipped with its own set of scores for each word. Finally, we describe and present examples of our generalized word shift graphs, showing how they create detailed summaries of how two texts differ.

As we have been doing, let us say that we have two texts $T^{(1)}$ and $T^{(2)}$ with relative word frequency distributions $P^{(1)}$ and $P^{(2)}$. Suppose, for now, that we have a single dictionary which assigns a score $\phi_\tau$ to each word $\tau$ in the vocabulary $\mathcal{T}$. Our main quantity of interest is the difference between the weighted averages $\Phi^{(1)}$ and $\Phi^{(2)}$,
\begin{equation}
    \Phi^{(2)} - \Phi^{(1)}
        =
            \sum_{\tau \in \mathcal{T}}
                \phi_\tau p_\tau^{(2)} 
             - 
             \sum_{\tau \in \mathcal{T}}
                \phi_\tau p_\tau^{(1)}.
    \nonumber
\label{eq:simple-weighted-avg}
\end{equation}
Denoting the difference as $\delta \Phi$, we can write it as the sum of contributions from each individual word,
\begin{equation}
    \delta \Phi
        =
            \sum_{\tau \in \mathcal{T}}
                \phi_\tau
                \bigl(
                    p_\tau^{(2)} - p_\tau^{(1)}
                \bigr)
        =
            \sum_{\tau \in \mathcal{T}}
                \delta \Phi_\tau,
\label{eq:sum-contributions}
\end{equation}
where we have introduced the notation $\delta \Phi_\tau$ for the summand.

To unpack the qualitatively different ways that words can contribute, we introduce $\phiref$, a \emph{reference score}. Consider the case of sentiment analysis. For each word in our score dictionary, we not only know its score, but also whether it is considered more or less positive. Importantly, the notion of being ``more'' or ``less'' positive is relative to some \emph{reference} value. For example, we may consider a word positive or not based on its position in an overall score distribution---words that are above the average score are positive and those that are below are not. Or instead, we may want to know which words make one text more positive than the other, in which case we can use the average sentiment of the reference text to determine which words are relatively positive. The quantity $\phiref$ encodes these kinds of reference points, distinguishing between different regimes of interest among word scores.

\begin{figure*}[t]
\centering
\includegraphics[scale=.30]{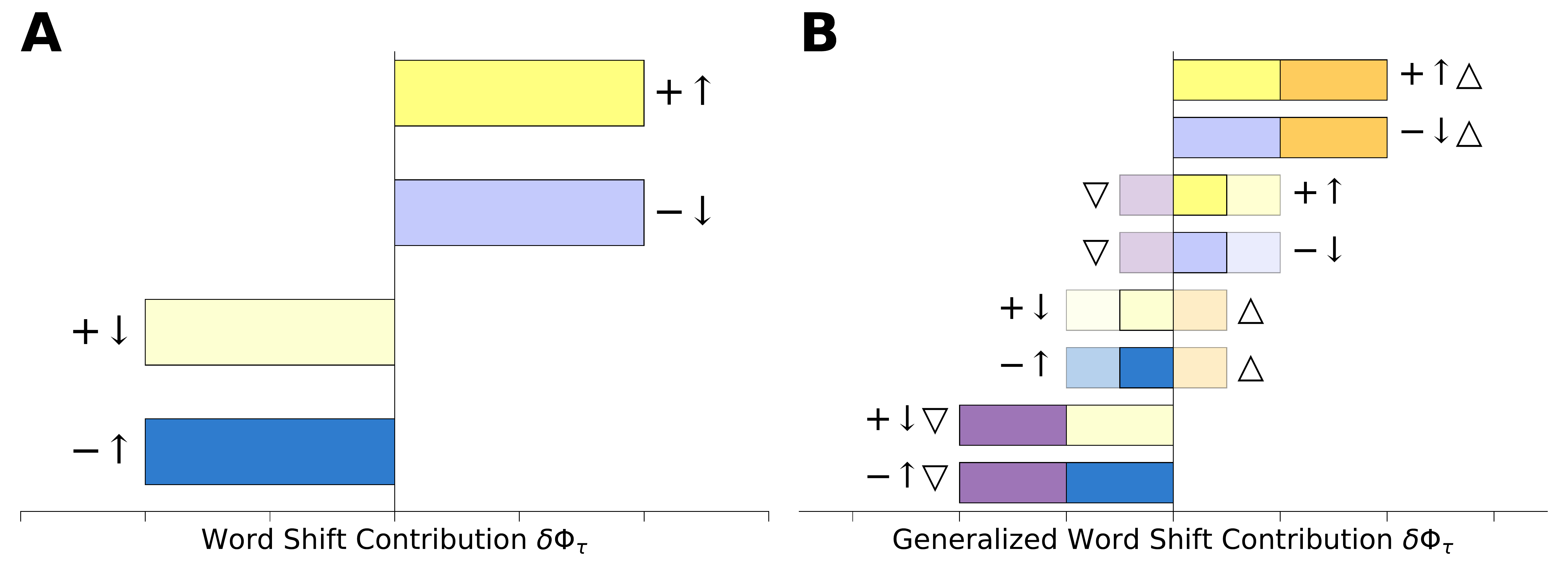}
\caption{
Types of word contributions in word shift graphs. \textbf{A)} Word contributions in basic word shift graphs, which are determined by the interaction between the signs of the difference between the word score and the reference score ($+ / -$) and the difference in relative frequencies ($\uparrow / \downarrow)$ (see Sec.~\ref{subsec.basicwordshifts}).
\textbf{B)} Word contributions in generalized word shift graphs, which additionally visualize the difference in word score ($\uptriangle / \downtriangle$) (see Sec.~\ref{subsec.generalizedwordshifts}). If component contributions counteract one another then they are faded to emphasize the magnitude of the resulting contribution while retaining information about the detraction of one component from the other
.
}
\label{fig:contributions}
\end{figure*}

Using the reference score $\phiref$, we can equivalently rewrite the sum of contributions (Eq.~\ref{eq:sum-contributions}) as
\begin{equation}
    \delta \Phi
        =
            \sum_{\tau \in \mathcal{T}}
            \bigl( p_\tau^{(2)} - p_\tau^{(1)} \bigr)
                \bigl( \phi_\tau - \phiref \bigr),
    \nonumber
\end{equation}
opening up a richer set of textual interpretations. Each word contribution is now the product of two components: the difference between the word score and the reference score, and the difference between relative frequencies. Both components can be either positive or negative, which yields four different ways that  a word can contribute, 
\begin{equation}
    \delta \Phi_\tau
        = 
            \underbrace{\bigl( p_\tau^{(2)} - p_\tau^{(1)} \bigr)}
                _{ \uparrow / \downarrow }
            \underbrace{\bigl( \phi_\tau - \phiref \bigr)}
                _{+ / -}.
    \nonumber
\end{equation}
If we say that $\phi_\tau  > \phiref$ implies that a score is ``relatively positive,'' and that $\phi_\tau < \phiref$ implies that a score is ``relatively negative,'' then without loss of generality we can colloquially phrase the ways that $T^{(2)}$ can have a higher score than $T^{(1)}$ as follows:
\begin{enumerate}
\item A relatively positive word ($+$) is used more often ($\uparrow$) in $T^{(2)}$ than in $T^{(1)}$.
\item A relatively negative word ($-$) is used less often ($\downarrow$) in $T^{(2)}$ than in $T^{(1)}$.
\end{enumerate}
Similarly, if $T^{(2)}$ has a higher score than $T^{(1)}$, two types of contributions counteract it to give $T^{(2)}$ a \emph{lower} score than it would have otherwise:
\begin{enumerate}
\item A relatively positive word ($+$) is used less often ($\downarrow$) in $T^{(2)}$ than in $T^{(1)}$.
\item A relatively negative word ($-$) is used more often ($\uparrow$) in $T^{(2)}$ than in $T^{(1)}$.
\end{enumerate}
While the language of ``positive'' and ``negative'' most conveniently maps onto the case of sentiment analysis, it is easily altered for other measures, e.g. a word may be ``relatively angry'' or ``relatively surprising'' if its score is larger than the reference score.

These contributions are the visual building blocks of word shift graphs (see Fig.~\ref{fig:contributions}A). If a word contribution is positive, $\delta \Phi_\tau > 0$ (i.e., $+\uparrow$ or $-\downarrow$), then the bar points to the right, and if it is negative, $\delta \Phi_\tau < 0$ (i.e. $+\downarrow$ or $-\uparrow$), then it points to the left. We use color and shading to differentiate the two different ways that each word can contribute in either direction. Relatively positive words ($+$) are colored in yellow and relatively negative words ($-$) are colored in blue, which is intuitive for sentiment word shifts, and colorblind friendly for any shift graph in general. Contributions that are due to an increase in word frequency ($\uparrow$) are shaded with deeper yellows and blues, while contributions from a decrease in word frequency ($\downarrow$) are shaded with lighter variations of the same colors. The direction, color, and shading succinctly summarize the four qualitatively different ways a word can contribute to the measurement variation between two texts.

\subsection{Generalized Word Shifts}
\label{subsec.generalizedwordshifts}

Already, we can start to see the richness that word shifts reveal. However, we also want to be able to account for words that have different scores in each corpus, such as with any of the entropy-based measures we introduced, or in sentiment analysis using domain-adapted score dictionaries \cite{hamilton2016inducing}. 

We introduce generalized word shifts, which allow words to take on corpus-specific weights. Rather than specifying a single score $\phi_\tau$ across both texts, let $\phi_\tau^{(i)}$ indicate that a word's score can be dependent on its appearance in either $T^{(1)}$ or $T^{(2)}$. The difference in weighted averages $\delta \Phi$ can then be written as
\begin{align}
\delta \Phi
    =
        \sum_{\tau \in \mathcal{T}}
           & \biggl(
                p_\tau^{(2)} - p_\tau^{(1)} 
            \biggr)
            \biggl[
                \frac{1}{2} 
                \bigl(
                    \phi_\tau^{(2)} + \phi_\tau^{(1)} 
                \bigr)
                - \phiref
            \biggr] \nonumber \\
            &+
            \frac{1}{2}
            \biggl(
                p_\tau^{(2)} + p_\tau^{(1)}
            \biggr)
            \biggl(
                \phi_\tau^{(2)} - \phi_\tau^{(1)}
            \bigg),
    \nonumber
\end{align}
where we provide full details of the derivation in the Materials and Methods. If the scores are the same, $\phi_\tau^{(1)} = \phi_\tau^{(2)}$, then we recover the basic word shift. When the word scores are, in fact, different, the average score of $\phi^{(1)}$ and $\phi^{(2)}$ is compared to the reference $\phiref$ to determine if the word is ``relatively positive'' or ``relatively negative.'' The second, new component in the generalized word shift accounts for the difference between the scores themselves, and weights it by the average frequency of the word. So in the generalized word shift framework, there are three major components to how a word contributes,
\begin{align}
\delta \Phi_\tau
    =
        &\overbrace{
            \biggl(
                 p_\tau^{(2)} - p_\tau^{(1)} 
            \biggr)
        }^{\uparrow / \downarrow}
        \overbrace{
            \biggl[
               \frac{1}{2} 
               \bigl(
                   \phi_\tau^{(2)} + \phi_\tau^{(1)} 
               \bigr)
                - \phiref
            \biggr] \nonumber
        }^{+ / -} \\
        &+
        \underbrace{
            \frac{1}{2}
            \biggl(
                p_\tau^{(2)} + p_\tau^{(1)}
            \biggr)
            \biggl(
                \phi_\tau^{(2)} - \phi_\tau^{(1)}
            \bigg)
        }_{\uptriangle / \downtriangle}.
    \label{eq:generalized-word-contribution}
\end{align}
This gives us eight distinct ways that a word can be visualized in a word shift graph (see Fig.~\ref{fig:contributions}B). Similar to before, we can visualize the interaction between the difference in relative frequency and the distance from the reference scores as yellow and blue bars. The new component, the difference between scores, is additive, which means that we can visualize it as an additional bar that augments or diminishes the base bar. When the signs of the two components of Eq.~\ref{eq:generalized-word-contribution} are congruent (as in the top two and bottom two bars of Fig.~\ref{fig:contributions}B), then we can visualize the score difference as an orange ($\uptriangle$) or purple ($\downtriangle$) stacked bar on top of the other. However, the two components can also counteract one another. In this case, each component's bar falls in a different direction, and we highlight this tension by coloring the contribution that remains after the counteraction, and fading the underlying offsetting components accordingly (as in the middle four bars). This maintains the full information about a particular word's components while emphasizing the word's overall contribution. The purple, orange, yellow, and blue bars are mutually colorblind friendly.

\subsection{Generalized Word Shift Graphs}

\begin{figure*}[hbt!]
\centering
    \subfloat{
        \includegraphics[scale=0.475]{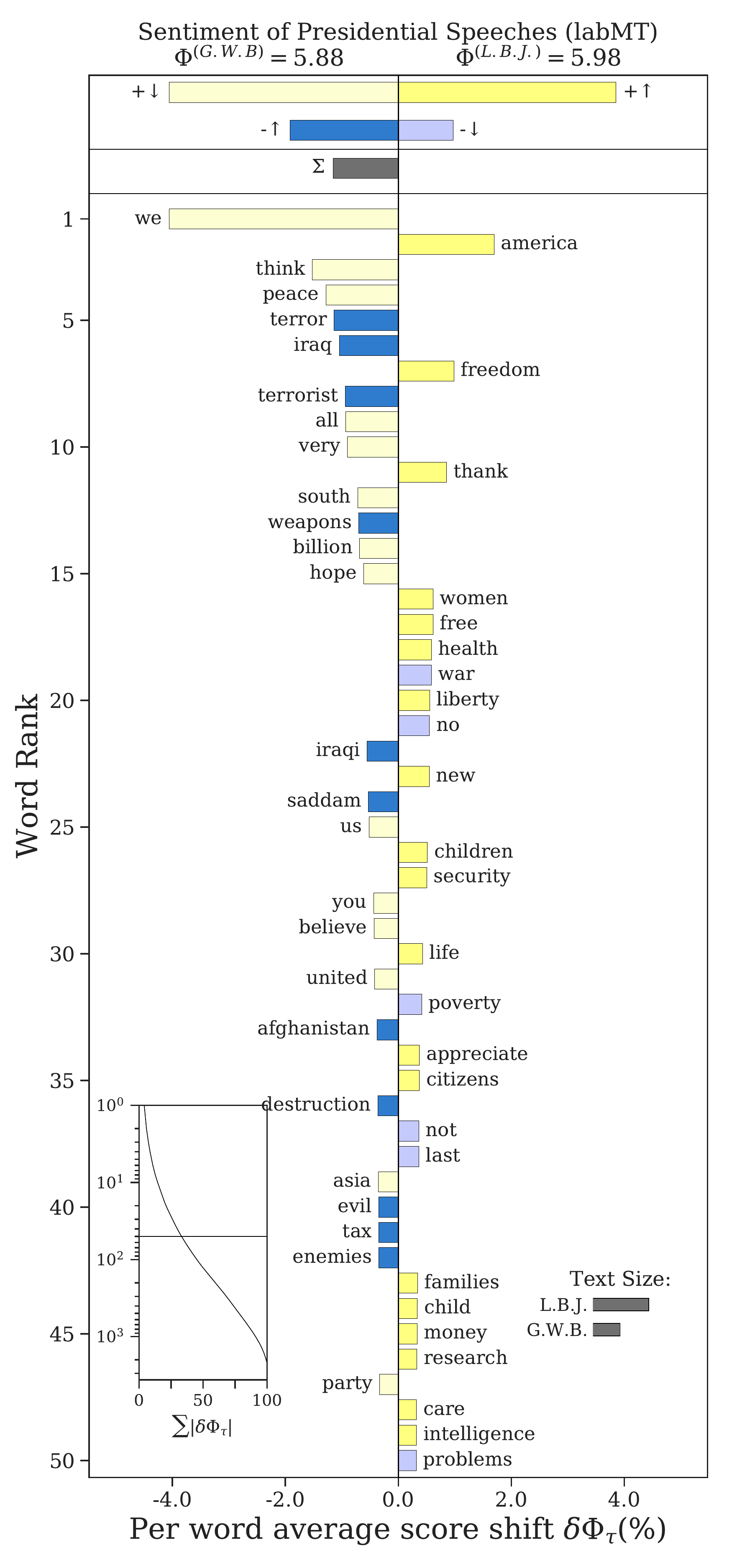}
    }
    \subfloat{
        \includegraphics[scale=0.475]{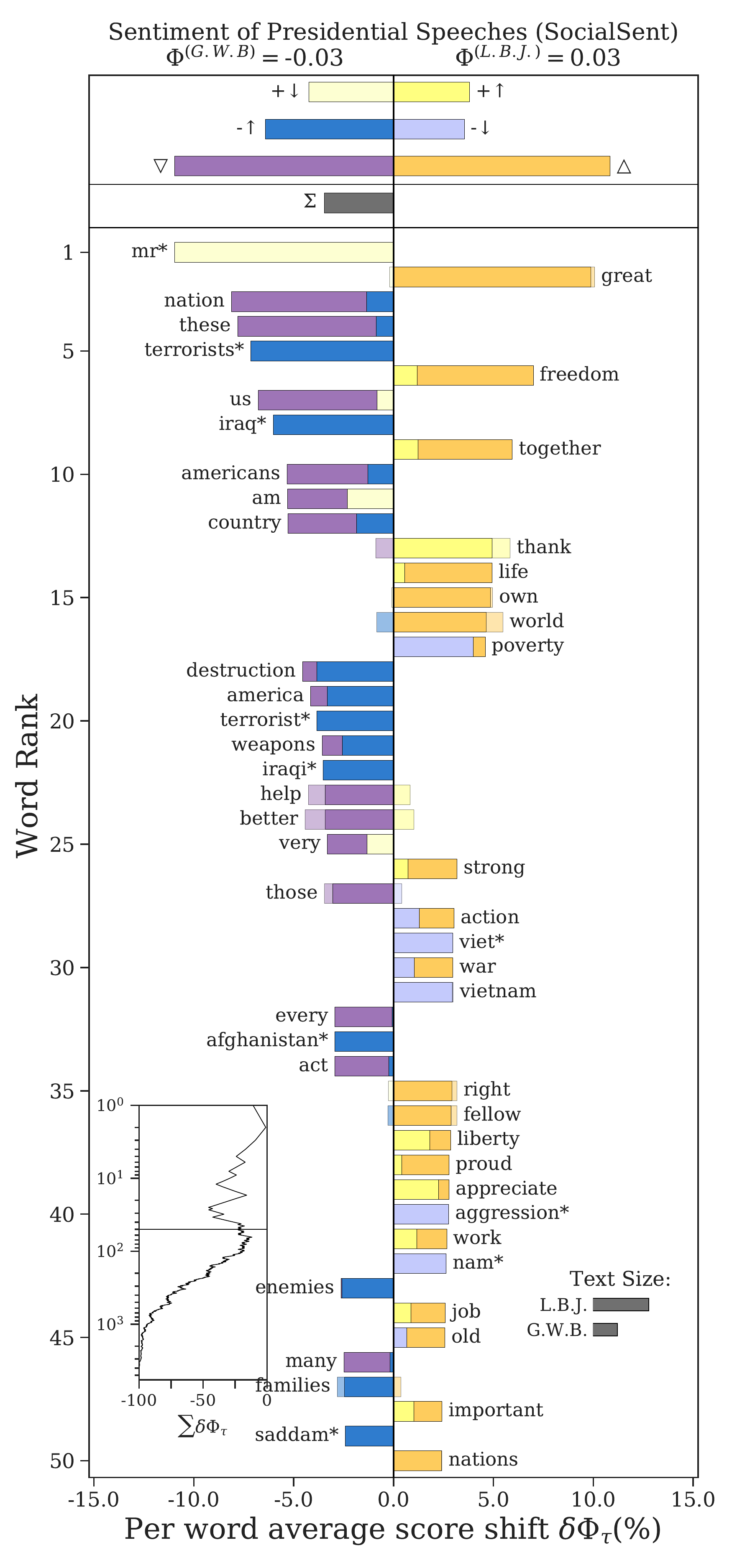}
    }
\caption{Word shift graphs of the sentiment of presidential speeches by United States Presidents Lyndon B. Johnson and George W. Bush. We display a basic word shift graph on the left, using the labMT  sentiment dictionary \cite{dodds2011temporal} (one score per word); on the right is a generalized word shift graph, using the SocialSent decade-adapted sentiment dictionaries \cite{hamilton2016inducing} (one score per word per decade). Word shifts show the top fifty contributing words to the difference in sentiment. Words to the left are those that contribute to Bush's speeches being more negative than Johnson's, while words to the right partly offset that negativity. These orientations are also reflected in the title, which displays the overall sentiment of each set of speeches. Bars at the top show the overall sentiment difference and the effect of each type of word contribution on that difference. The inset in the bottom left shows how the word shift scores cumulatively vary as a function of word rank, where the horizontal line in the middle of the plot indicates what is explained by the top fifty words shown in the graph. The inset in the bottom right shows the proportional size of each corpus with respect to number of word tokens. In the generalized word shift graph, words that are borrowing a score across decades are marked by an asterisk ($\ast$). We use the center of each of labMT's and SocialSent's sentiment distributions as the reference scores: $\Phi^{(\text{ref})} = 5$ for labMT and $\Phi^{(\text{ref})} = 0$ for SocialSent.}
\label{fig:word-shift-examples}
\end{figure*}

We now present generalized word shift graphs in their entirety. For our visual case study, we compare the average sentiment of speeches by two United States presidents: Lyndon B. Johnson (1963 -- 1969) and George W. Bush (2001 -- 2009). We use the labMT sentiment dictionary \cite{dodds2011temporal} to construct a basic word shift, and the SocialSent historical sentiment lexicons \cite{hamilton2016inducing} for a generalized word shift. The SocialSent lexicons are decade-specific sentiment dictionaries that were adapted for each decade between 1850 and 2000 by using semi-supervised machine learning and the Corpus of Historical American English, and so words can take on different scores depending on what sentiment they were associated with in the 1960s or 2000s. We use the word shift graphs (presented in Fig.~\ref{fig:word-shift-examples}) primarily as visual examples, and so we focus more on their construction and layout rather than their substantive interpretation.

\subsubsection{Word Contributions}

We measure the difference in average sentiment of the presidential speeches, $\Phi^{(G.W.B.)} - \Phi^{(L.B.J.)}$, and rank words by their absolute contribution to that difference. According to both dictionaries that we have employed, Bush's speeches were more negative than Johnson's, as indicated by the average sentiments displayed in the title of each graph. We plot word contributions as a horizontal bar chart, where words that contribute to the negativity of Bush's speeches are directed to the left, while words that counteract $\Phi^{(G.W.B.)} < \Phi^{(L.B.J.)}$ point to the right.

Examining the basic word shift on the left, we see that Bush's use of more negative words ($-\uparrow$), like `terror', `weapons', and `tax', all lower the sentiment of his speech's relative to Johnson. Further, the decreased use of positive words ($+\downarrow$), such as `we', `peace', and `hope', also contributes to the negativity of Bush's speech. On the other hand, these contributions are partly offset by a lesser use of negative words $(-\downarrow)$ like `no', `poverty', and `problems', and greater use of positive words $(+\uparrow)$ like `america', and `freedom'. 

In the generalized word shift to the right, we see that changes in the sentiments of the words themselves also affect the overall difference between Bush's and Johnson's speeches. The words `nation', `us', and `destruction' are all associated with more negativity ($\downtriangle$) in the 2000s than in the 1960s. Similarly, but in the opposite direction, `freedom', `together', and `life', are all associated with more positivity ($\uptriangle$) in the 2000s than the 1920s. We also see counteracting contributions for individual words: `better', for example, is a positive word that was used more by Bush, but its positive contribution is offset by its decline in sentiment from the 1960s to the 2000s.

We mark some words with asterisks ($\ast$) to indicate that they have ``borrowed'' a score across decades. For example, `iraqi' is not associated with a sentiment in the 1960s dictionary, while `viet' is not associated with a sentiment in the 2000s dictionary. This can happen when sentiment ratings are prioritized by word frequency to maximize coverage (see Materials and Methods for details).

{\color{white} blank text blank text blank text blank text blank text blank text blank text blank text blank text blank text} Overall, the generalized word shift graph succinctly visualizes which words contribute to the negativity of George W. Bush's speeches relative to Lyndon B. Johnson and, importantly, how they do so. The word shift graphs distinguish between subtle differences in contributions, such as whether the speeches are more negative because more negative words were used or less positive ones were. Rather than just comparing two averages, like $\Phi^{(G.W.B.)} = -0.03$ and $\Phi^{(L.B.J.)} = 0.03$, the word shift graphs allow us to simultaneously quantify word usage, sentiment, and temporal drift to tell a richer story about how, plausibly, Bush's speeches were negative in part due to their focus on the Iraq War starting in 2003 and, perhaps, also in part due to decreased positivity associated with nationalistic words like `nation', `us', `country', `america', and `americans'.

\begingroup
\renewcommand\arraystretch{3.0}
\setlength{\tabcolsep}{10pt}
\begin{table*}[!htb]
\begin{tabular}{lll}
\textbf{Measure}
    & \textbf{Notation}  
    & \textbf{Word Contribution} $\delta \Phi_\tau = p_\tau^{(2)}\phi_\tau^{(2)} - p_\tau^{(1)} \phi_\tau^{(1)}$  
\\
\hline
Relative Frequency 
    & $P^{(i)}$ 
    & $p_\tau^{(2)} - p_\tau^{(1)}$ 
\\
\hline
Shannon Entropy 
    & $H\bigl(P^{(i)}\bigr)$  
    & $- p_\tau^{(2)} \log p_\tau^{(2)} + p_\tau^{(1)} \log p_\tau^{(1)}$ 
\\
\hline
Generalized Entropy 
    & $H_\alpha\bigl(P^{(i)}\bigr)$ 
    & $-p_\tau^{(2)} \biggl[ \frac{\bigl(p_\tau^{(2)}\bigr)^{\alpha-1}}{\alpha-1} \biggr]
        + p_\tau^{(1)} \biggl[ \frac{\bigl(p_\tau^{(1)}\bigr)^{\alpha-1}}{\alpha-1} \biggr]$
\\
\hline
Kullback-Leibler Divergence 
    & $D^{(\text{KL})}\bigl(P^{(2)} || P^{(1)}\bigr)$ 
    & $- p_\tau^{(2)} \log p_\tau^{(1)} + p_\tau^{(2)} \log p_\tau^{(2)}$ 
\\
\hline
Jensen-Shannon Divergence  
    & $D^{(\text{JS})}\bigl(P^{(1)} || P^{(2)}\bigr)$  
    & $p_\tau^{(2)} \pi_2 \log \frac{p_\tau^{(2)}}{m_\tau}
        - p_\tau^{(1)} \pi_1 \log \frac{m_\tau}{p_\tau^{(1)}}$
\\
\hline
Generalized Jensen-Shannon Divergence
    & $D_\alpha^{(\text{JS})}\bigl(P^{(1)} || P^{(2)}\bigr)$  
    & $p_\tau^{(2)} \pi_2 \biggl[ \frac{\bigl(p_\tau^{(2)} \bigr)^{\alpha-1} - m_\tau^{\alpha-1}}{\alpha-1} \biggr]
        - p_\tau^{(1)} \pi_1 \biggl[\frac{m_\tau^{\alpha-1} - \bigl(p_\tau^{(1)}\bigr)^{\alpha-1}}{\alpha-1}\biggr]$
\end{tabular}
\caption{Contributions and scores of various text comparison measures according to the word shift framework. The word contribution $\delta \Phi_\tau$ indicates how an individual word impacts a measure, and each contribution is expressed as a difference in weighted averages so that it can be easily identified with the components of the word shift framework.}
\label{tab:word-shifts}
\end{table*}
\endgroup

\subsubsection{Cumulative Contributions}

Along with the bar chart of word contributions, we also visualize several higher level summary quantities that help us better contextualize the individual word dynamics. In the bottom right corner of the word shift graphs, we visualize the relative size of the two corpora---which shows that Johnson's speech corpus is twice the size of Bush's. At the top of the figures, we display how each distinct type of word shift contributes to the total difference, $\Sigma$. In the basic word shift graph, we see that the negativity of Bush's speeches is most explained overall by the use of more negative words ($-\uparrow$) and less positive words ($+\downarrow$). In the generalized word shift, the negativity is most affected by the general negative shift in word sentiment ($\downtriangle$) from the 1960s to the 2000s, though that component is largely offset by other words increasing in sentiment ($\uptriangle$). These summary totals help accumulate sentiment information across all of the words and tell us what qualitative factors play the largest roles in differentiating the speeches of Presidents Bush and Johnson.

The inset in the bottom left corner of the word shift graphs illustrates how each word cumulatively impacts the sentiment difference. This diagnostic plot shows how the difference $\delta \Phi$ changes as we add words according to their rank, where the horizontal line demarcates the boundary between the top fifty words we see in the plot and the thousands of others in the presidential speeches. There are two ways we can visualize the cumulative contributions. The first, shown in the basic word shift on the left, plots the normalized absolute magnitude of the contributions,
\begin{equation}
    \sum_{\tau \in \mathcal{T}}
        | \delta \Phi_\tau |,
    \nonumber
\end{equation}
which measures what percent of all variation is explained up to a given word rank. For example, a bit more than a quarter of the variation in sentiment between Lyndon B. Johnson's and George W. Bush's speeches is explained by the top fifty words. The second way to cumulatively visualize contributions is shown on the right in the generalized word shift graph, where we plot $\sum_\tau \delta \Phi_\tau$ as a function of rank and normalize by
\begin{equation}
    | \delta \Phi |
        =
            \biggl|
                \sum_{\tau \in \mathcal{T}}
                    \delta \Phi_\tau
            \biggr|.
    \nonumber
\end{equation}
This displays the trajectory of the sentiment difference as we add additional words, which helps highlight effects of the distributional tail that are not apparent among the top fifty words. Together, the inset cumulative rank contribution plot and the total contribution bars give us important summaries of how the individual word contributions come together in total, and they draw our attention to textual differences that may not be explained by the high ranking words that are visualized in the bar chart.

\subsection{Pairwise Comparison Measures as Word Shifts}

We have shown how dictionary scores can be naturally incorporated into the word shift framework. We now return to the other text comparison measures that we introduced earlier: relative frequency, Shannon entropy, the Kullback-Leibler divergence (KLD), the Jensen-Shannon divergence (JSD), and their generalized forms (see Materials and Methods for details). Some of these measures, like relative frequency and the Shannon entropy, are easily identifiable as weighted averages by how they are commonly written. Other measures though, like the KLD, the JSD, and the generalized entropies, can often be expressed in ways that do not make it clear that they are also weighted averages. In Table~\ref{tab:word-shifts}, we explicitly write the word contribution $\delta \Phi_\tau$ of each measure as a difference in weighted averages. Making this form explicit allows us to easily situate all of these measures within the generalized word shift framework and visualize them through generalized word shift graphs.

Formulating each measure in terms of weighted averages is one of two key elements for using generalized word shift graphs. The other is identifying a reference score $\phiref$ that discerns between distinct and interesting regimes of the word scores. As we have seen with sentiment analysis, one obvious candidate for the reference score is the center of the sentiment scale, which naturally sifts positive words from negative ones. While, in practice, researchers rarely draw an explicit boundary between different types of words when using the measures presented in Table~\ref{tab:word-shifts}, the generalized word shift framework provides us an opportunity to be more intentional and creative with how we quantify, interpret, and visualize differences between texts. For example, researchers often remove commonly appearing ``stop words'' \cite{grimmer2013text,denny2018text} by applying a pre-assembled list or identifying the top, say, 1\% of frequently occurring words. Rather than discarding them in an ad hoc manner though, we may instead choose to leave them in and use them to help us mark the boundary between frequent and infrequent, surprising and unsurprising. Or if we are working with an entropy-based measure, we may set the reference value to be the average entropy of one of the texts. Using the text's entropy as a reference allows us to discern which words contribute to a text's unpredictability because they are even more surprising than the average surprisal.

Of course, it is always mathematically valid to set $\phiref = 0$ if we are satisfied with just knowing \emph{which} words distinguish two texts. However, doing so always risks masking the richness in \emph{how} those words contribute. Placing the frequency and entropy-based measures within the generalized word shift framework gives us new ways of understanding them and disentangling the complexities of text as data.

\section{Case Studies: Using Word Shift Graphs in Practice}

To show how generalized word shift graphs can be used in practice, we present a diverse set of case studies that highlight how they can be used as both a diagnostic tool during the research process and an illustrative instrument for scientific communication. First, through sentiment analyses of both the book \emph{Moby Dick} and U.S. urban parks, we demonstrate how word shifts warn us when there are significant measurement issues that require us to revisit how the text is preprocessed and quantified. Second, through a case study of Twitter's change from 140 to 280 character tweets, we show how word shifts make it possible to interpret unexplained textual trends and generate additional research hypotheses. Finally, through a case study of labor diversity and the Great Recession, we show how shift graphs enrich analyses beyond just the research process and provide fine-grained evidence that support deeper substantive insights by domain experts.

\subsection{Sentiment Peculiarities of Moby Dick and U.S. Urban Parks}

\begin{figure*}[hbt!]
\centering
    \subfloat{
        \includegraphics[scale=0.475]{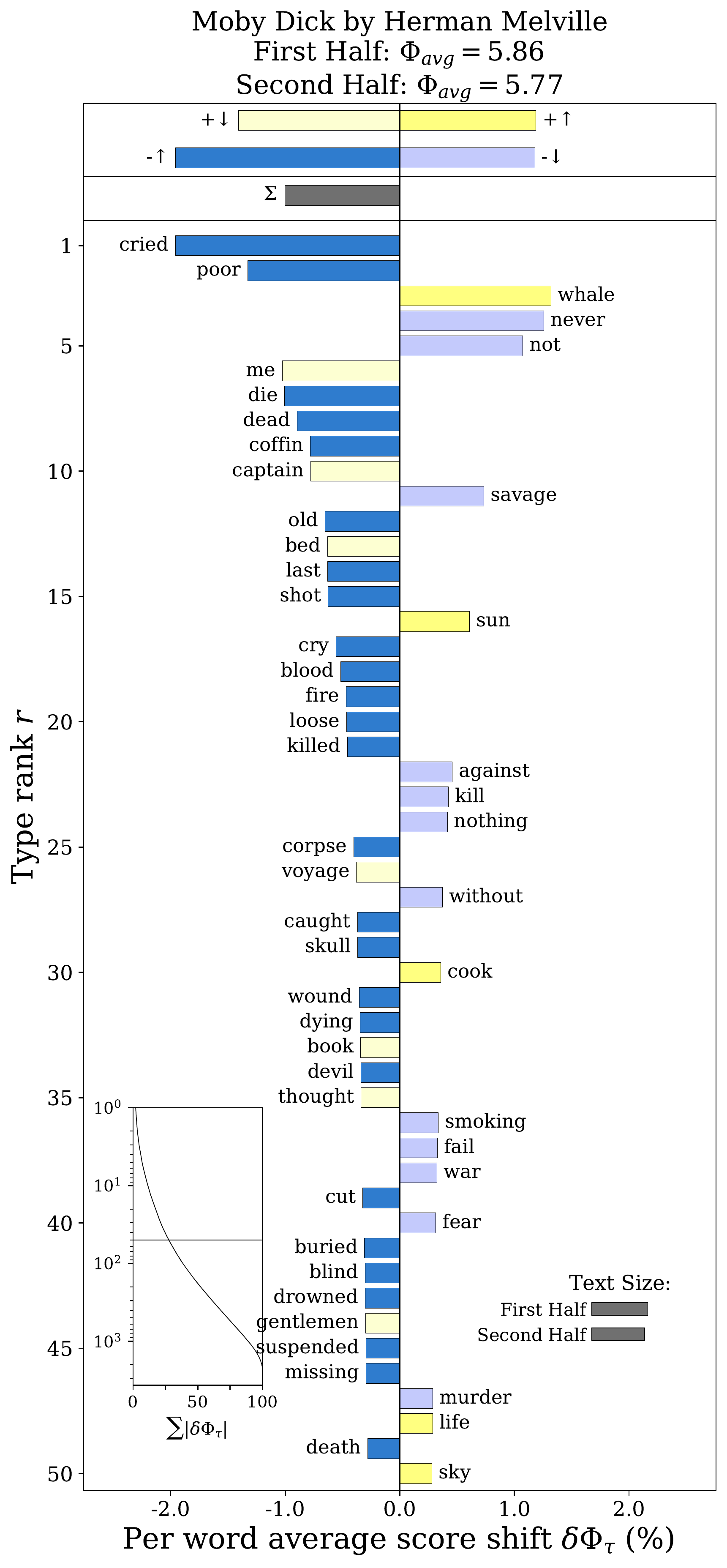}
    }
    \subfloat{
        \includegraphics[scale=0.475]{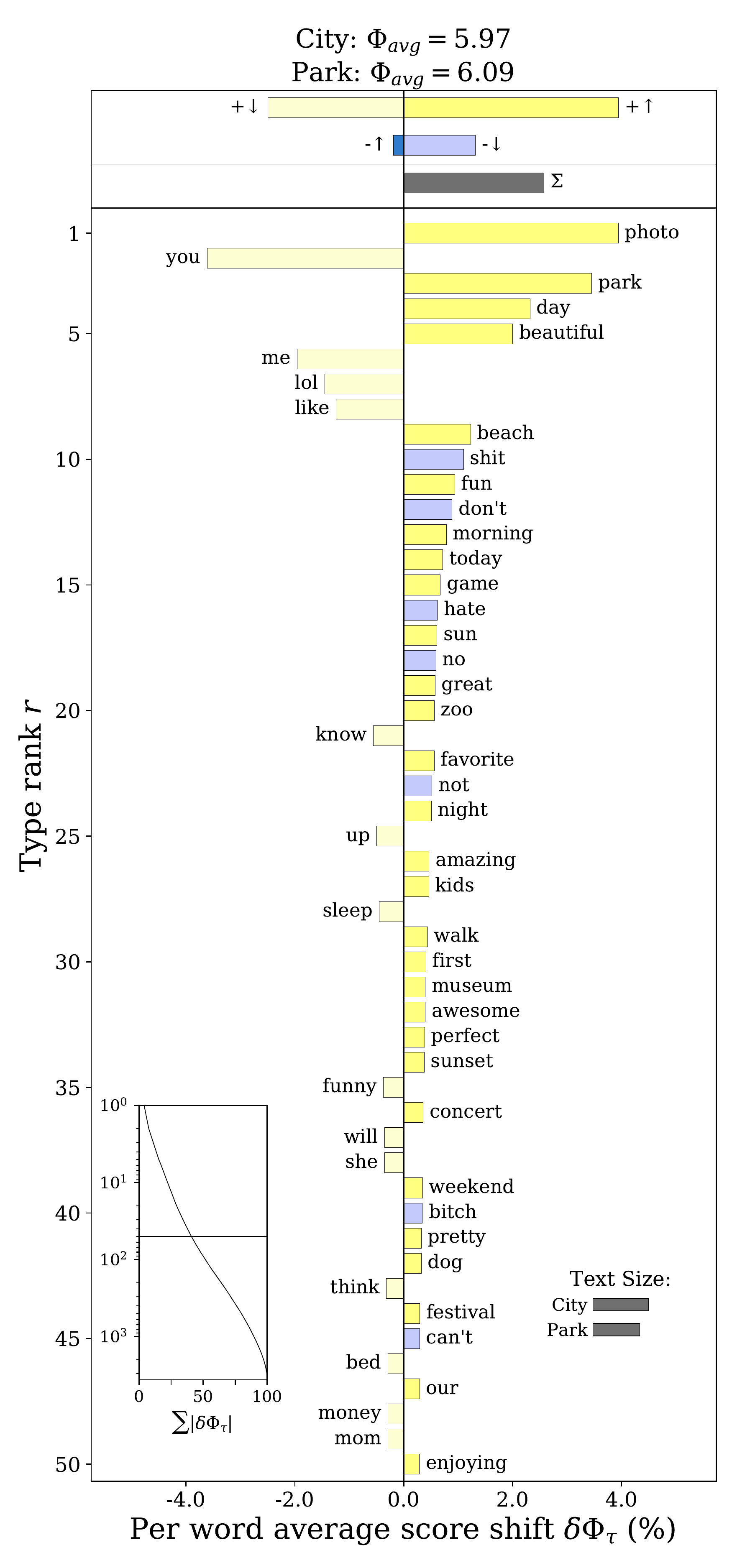}
    }
\caption{\textbf{Left)} Word shift graph of the sentiment difference between the first and second halves of \emph{Moby Dick} by Herman Melville. A naive application of a dictionary-based sentiment lexicon for the two-segment emotional arc would inflate the negative trajectory of the novel without the preprocessing or removal of words like `cried' and `cry', which more often means `said' in nineteenth century English, and `coffin', which is used as a surname about one third of the time. We find that $\delta \Phi = 0.09$  when including those words, while $\delta \Phi = 0.07$ when they are excluded. \textbf{Right)} Word shift graph of the sentiment difference between the in-park and out-of-park tweets across 25 cities in the US. A naive application of a dictionary-based sentiment lexicon would inflate the in-park tweet scores by including words like `park', `beach', `zoo', `museum', `music', and `festival', all of which represent physical locations and events within parks. We find that $\delta \Phi = 0.12$  when including those words, while $\delta \Phi = 0.10$ when they are excluded. For both word shift graphs, a reference value of 5 was used, and a stop lens was applied on all words with a sentiment score between 4 and 6.}
\label{fig:mismeasurement}
\end{figure*}

Dictionary-based sentiment analysis is sensitive, of course, to the dictionary that is used. Sentiment dictionaries are often static objects, constructed once for general use. This can be problematic if there has been a temporal shift in how particular words or used, or when words take on different sentiments in particular contexts \cite{loughran2011liability,grimmer2013text}. As we show, word shift graphs transparently diagnose these kinds of measurement issues.

We start with a case study of \emph{Moby Dick}, the 1851 novel by Herman Melville. We naively apply the labMT sentiment dictionary \cite{dodds2011temporal} to the first and second halves of the book, a simple quantification of the novel's emotional arc \cite{reagan2016emotional}. The sentiment word shift graph is shown on the left in Fig.~\ref{fig:mismeasurement}. There are two issues that are made visible by the word shift graph, each of which we could easily miss otherwise. First, examining the left panel of Fig.~\ref{fig:mismeasurement}, the overall sentiment is affected considerably by the words `cried' and `cry.' Throughout the book though, `cried' and `cry' are often understood to mean `said'. Second, the word `coffin' also significantly affects the sentiment. However, while coffins are mentioned throughout the story, searching the raw text of the novel\footnote{The text of \emph{Moby Dick} is freely available at \url{http://www.gutenberg.org/files/2701/2701.txt}} reveals that about a third of its usage is with respect to the surname `Coffin.' All three of these words contribute in an unintended way to the sentiment differences between the first and second half the book---with them included, the difference in sentiment is $\delta \Phi = 0.09$, while without them it is $\delta \Phi = 0.07$, a difference of over 20\%.

Word shift graphs make these contributions apparent. One way to address these issues is through additional text preprocessing. For example, removing only capitalized uses of `Coffin' (along with `cry' and `cried') allows for `coffin' to still contribute, yielding a sentiment difference of $\delta \Phi = 0.08$, which is 15\% less than the naive approach. Another way is through modification of the dictionary itself---domain knowledge or semi-supervised machine learning \cite{hamilton2016inducing} can help refine or adapt the sentiment dictionary to the language of nineteenth century English. By highlighting these mismeasurements early in the research process, word shift graphs allow researchers to make appropriate adjustments in the data pipeline.

To emphasize the need for word shift graphs in identifying bias induced by sentiment mismeasurement, we also consider a case study of tweets posted inside and outside of U.S. urban parks. Prior work has demonstrated that people are happier when visiting urban green spaces such as parks \cite{schwartz2019visitors}, and social media data presents an opportunity to supplement traditional survey measures with geographically fine-grained measurements. However, naively applying the labMT sentiment dictionary to tweets may overestimate the sentiment difference between in- and out-of-park tweets. In the right panel of Fig~\ref{fig:mismeasurement}, we see that the word `park' is contributing substantially to the higher sentiment of in-park tweets. However, in the context of inferring happiness from tweets, writing the word `park' is often simply a declaration of where a user is located, rather than a proxy for how they may be feeling. Similarly, words like `museum', `zoo', and `beach' also represent physical locations within parks, but contribute to the positivity of in-park tweets because they are all relatively positive words. `Music' and `festival' also appear frequently within park tweets, which are related to events in parks, but often not nature itself. 

While there are defensible arguments for and against removing each of these words, word shift graphs make their contributions visible, and allow a researcher to make transparent decisions with the understanding of how results may change based on which words are included in the final analysis. When removing the above six words, the sentiment difference goes from $\delta \Phi = 0.12$ to $\delta \Phi = 0.10$, more than a 15\% difference. Adjustments for specific words, in tandem with the examination of a word shift graph, allow us to apply sentiment analysis with the confidence that one or a few individual words have not made a folly of our analyses.

\subsection{Information Content of 280 Character Tweets}

\begin{figure}[tp!]
\centering
\includegraphics[width=\columnwidth]{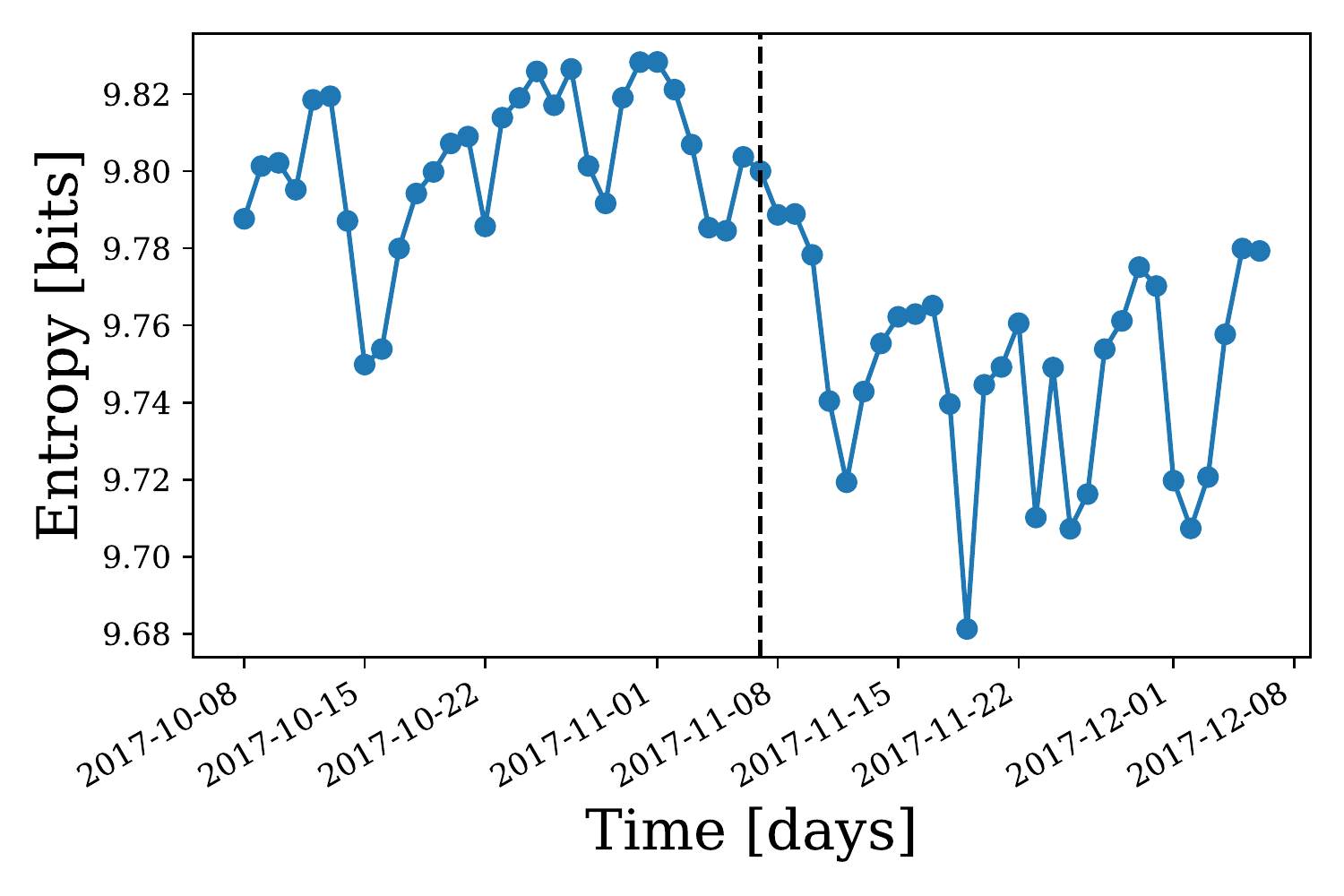}
\caption{Time series indicating the change in the entropy of Twitter language before and after the platform's change from 140 character to 280 character tweets (marked by dashed line).}
\label{fig:daily_entropy}
\end{figure}

On November 7th, 2017, Twitter doubled the character limit for all tweets from 140 to 280 characters, one of the most significant changes to the platform since its inception in 2006. Prior to the change, Twitter found that there were discrepancies in how often users reached the 140 character limit based on the language in which they tweeted \cite{ihara2017our}. They partly attributed the discrepancies to the ability of different languages to encode more or less information in a single character \cite{ihara2017our,neubig2013much}; for example, users tweeting English hit the limit 9\% of the time, while those tweeting in Japanese rarely did so. Immediately following the update to 280 character tweets, English users only hit the limit about 1\% of the time, suggesting that the change made it ``easier to tweet'' by making it easier for individuals to spend ``less time editing their tweets in the composer'' \cite{Rosen2017}. Outside of Twitter, it has been independently verified that the increased character limit increased the Flesch-Kincaid reading level of tweets and decreased the proportion of users hitting the character limit \cite{mitchell2018mo}.

\begin{figure}[tp!]
\centering
\includegraphics[width=\columnwidth]{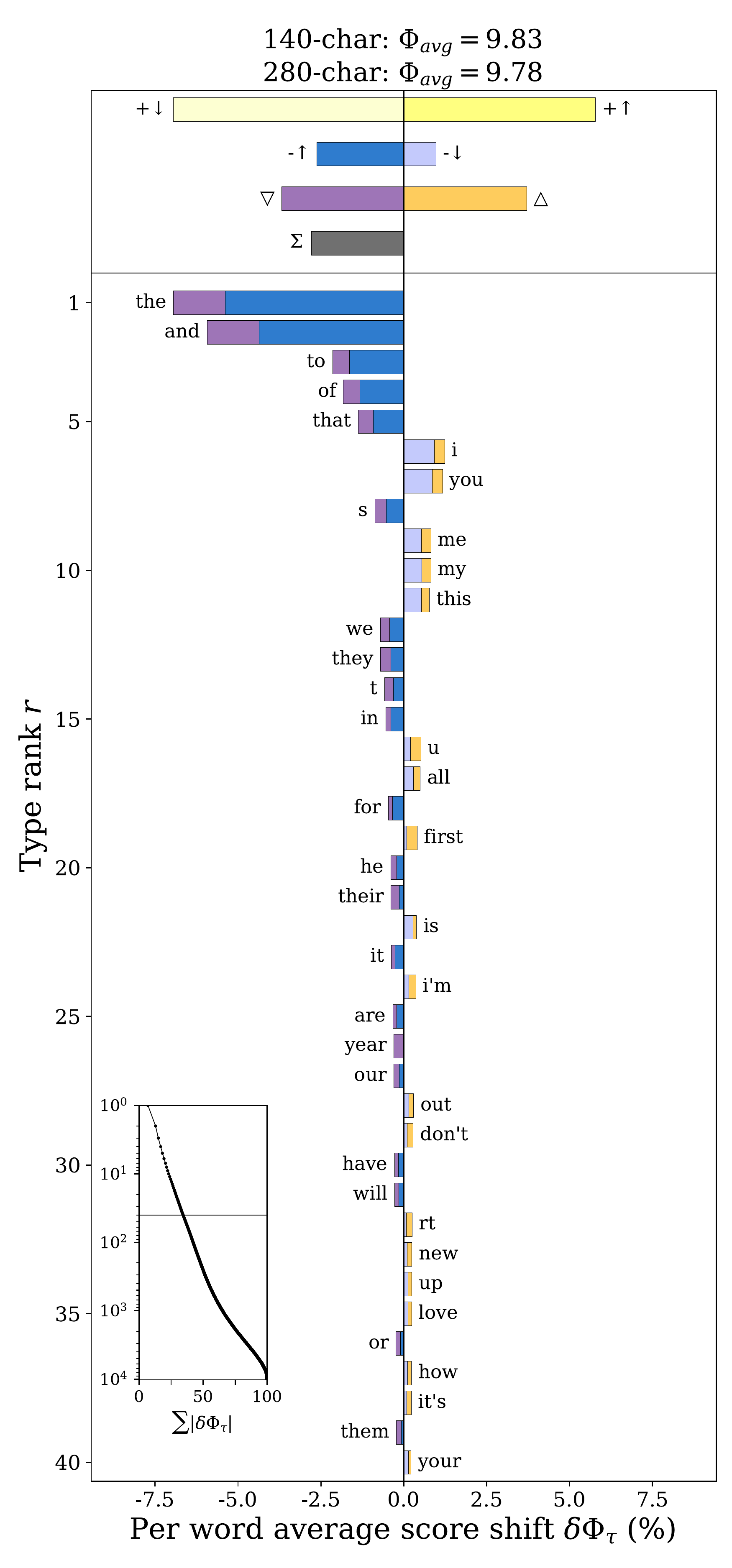}
\caption{Generalized word shift graph of the change in Shannon entropy for the 30 days before and after the 140 to 280 character change on Twitter. Words are relatively ``surprising'' (+) or ``unsurprising'' (-) depending on if their surprisal is higher than the entropy, or average surprisal, of words used in 140 character tweets. Many of the contributions in the top forty words are from unsurprising words being used relatively more ($-\uparrow$) or less $(-\downarrow)$. The surprisal for each word went down ($\downtriangle$) or up ($\uptriangle$) depending on if it was used more or less, respectively, in 280 character tweets. As seen by the top of the word shift graph and the cumulative inset, most contributions come from a long tail of relatively surprising words, despite mostly not appearing among the top forty words.
}
\label{fig:entropy_shift}
\end{figure}

Moving from individual characters to whole words, we measure the Shannon entropy of the Twitter word distribution before and after the 280 character change to understand how the character-level change may have affected the information content of the language used in tweets. 
We take all tweets collected from Twitter's Decahose, a 10\% sample of tweets, 30 days before and after the change and aggregate them into two separate bags of words, using the labMT dictionary as our vocabulary (see Materials and Methods for details). Fig.~\ref{fig:daily_entropy} shows that the information content of tweet language decreased after the change to 280 characters. 
Using a generalized word shift graph, we can reveal what words specifically contributed to that drop and why (see Fig.~\ref{fig:entropy_shift}). We use the entropy before the change to 280 character tweets as our reference value, implying that a word is considered relatively ``surprising'' if its surprisal is higher than the average word surprisal in 140 character tweets.

As we may expect from a change allowing for longer tweets, the top five contributions to the decrease in entropy all come from greater use of the most common parts of speech, including conjunctions (`and', `that'), articles (`the') and prepositions (`to', `of'). All of these are relatively ``unsurprising'' words, so they drive the entropy down in 280 character tweets. It is plausible that in the shift to longer tweets, users are able to write longer messages and use less abbreviations, allowing them to place a heavier on traditional function words. Further, the entropy word shift also reveals some more unexpected trends, namely a decrease in first- and second-person personal pronouns (`i', `you', `me', `my', `u', `i'm' and `your') and an increase in third-person pronouns (`we', `they', `their', `our', and `them').
This is somewhat striking, particularly as it is an observation that has emerged from the data in an unsupervised manner. Finally, note that the majority of the top forty contributions are from relatively unsurprising words. As shown by the corner inset, these explain a bit more than 30\% of the total entropy difference between 140 character language and 280 character language. Yet, by the top of the word shift graph, we see that the largest contributions come from the use of relatively surprising words, few of which appear in our figure. This suggests that there is a richer story in the long tail of the word distribution, which would not be obvious without the word shift graph diagnostics.

Through a brief investigation of the change from 140 to 280 character tweets, generalized word shift graphs have allowed us to uncover three potentially fruitful hypotheses: Twitter users do not need to abbreviate common function words as often, tweets deploy more collective framing through third-person pronouns, and less common words account for the largest shift in entropy. Of course, all of these hypotheses are speculative and require much deeper investigations. This is exactly what demonstrates the power of word shift graphs though. These stories are hidden by the aggregate entropy measures, which obscure why the entropy dropped after the character limit change. Generalized word shift graphs unpack these measures and allow us to quickly generate new questions and hypotheses that bring our research in directions that may have been otherwise unexplored.

\subsection{Employment Diversity and Urban Resilience during the Great Recession}

\begin{figure}[tp!]
\centering
\includegraphics[width=\columnwidth]{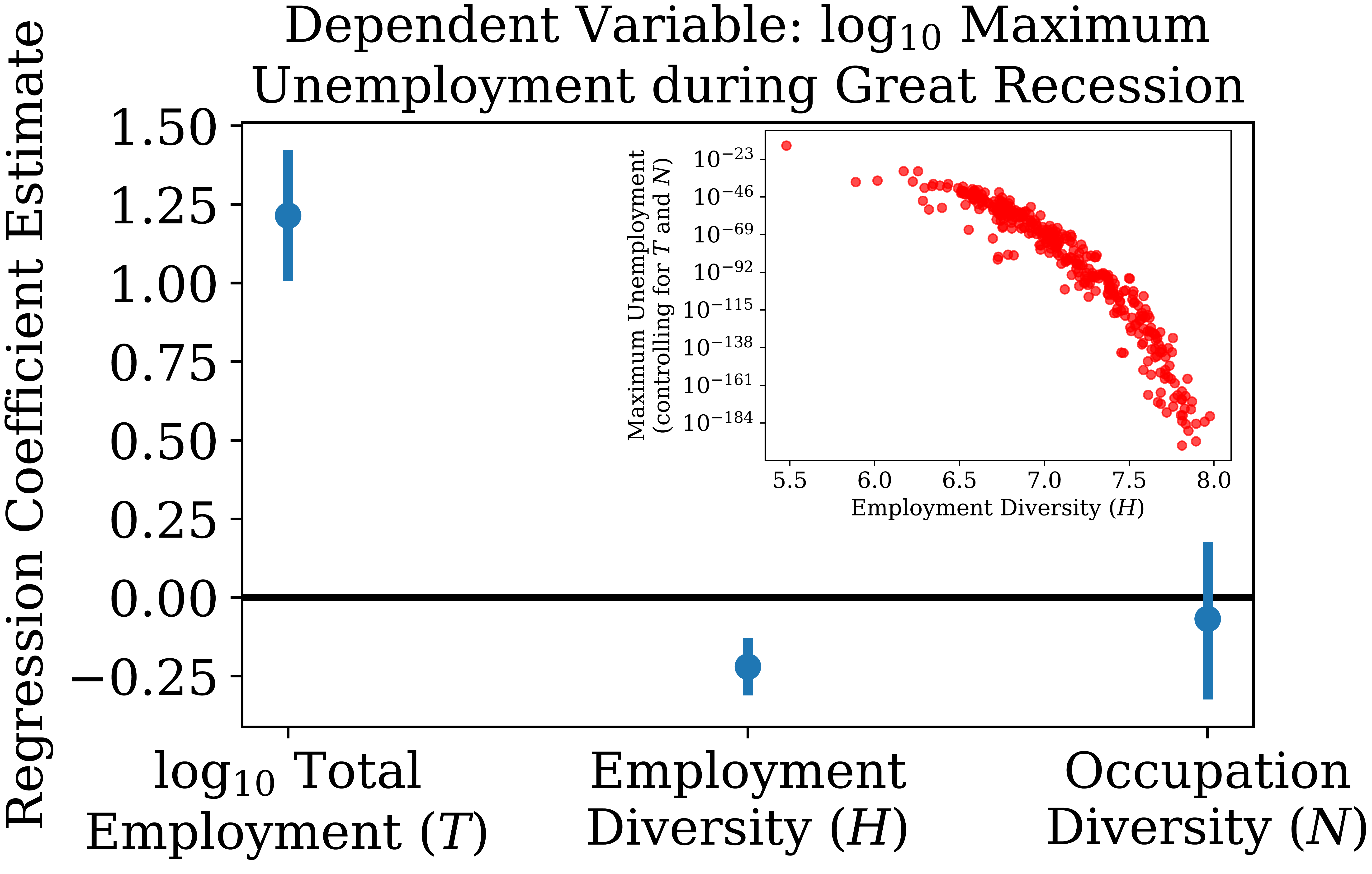}
\caption{
    Linear regression coefficients showing that U.S. cities with greater employment diversity $H^{(c)}$ saw lower peak unemployment during the Great Recession, after controlling for total employment $T^{(c)}$ and the number of unique occupations in each city $N^{(c)}$.
    All variables were centered and standardized prior to regression.
    Error bars represent 95\% confidence intervals.
    Additional details on the regression analysis are provided in the Materials and Methods.
}
\label{fig:urbanResilience}
\end{figure}

The Great Recession, which spanned from the end of 2007 to 2012, is one of the most significant economic disruptions in the United States' history \cite{elsby2010labor}. Understanding which U.S. cities were more or less resilient to the the recession and why could inform urban policy that diminishes the disruptions to labor and employment. Similar to ecological systems \cite{oliver2015biodiversity}, and complex systems more generally, employment diversity is hypothesized to play a key role in the resilience of urban labor markets. Diverse labor markets have more potential for redundancies among occupations that enable local workers and firms to adapt to disruptions.
For example, one recent study compared labor and skill diversity in cities to the cities' exposure to automation~\cite{frank2018small} and found that labor market diversity was more predictive than the size or regional economy of the city.

To study urban response to the recession, we turn to the U.S. Bureau of Labor Statistics (BLS), which records employment data for cities across the United States. If we consider each city to be a ``corpus'' and each distinct occupation to be a ``word,'' then we can use the word shift framework to understand differences in employment diversity between cities. Let $\mathcal{J}$ be the ``vocabulary'' of the 794 jobs recorded by the BLS in 2007 across 375 U.S. cities, and denote the number of people employed with job $j$ in city $c$ as $f_j^{(c)}$. The total employment across the entire urban labor market is $T^{(c)} = \sum_{j \in \mathcal{J}^{(c)}}f_j^{(c)}$, and the relative frequency of a job in the labor distribution $P^{(c)}$ is $p_j^{(c)} = f_j^{(c)} / T^{(c)}$, like our word distributions.

\begin{figure}[tp!]
\centering
\includegraphics[width=\columnwidth]{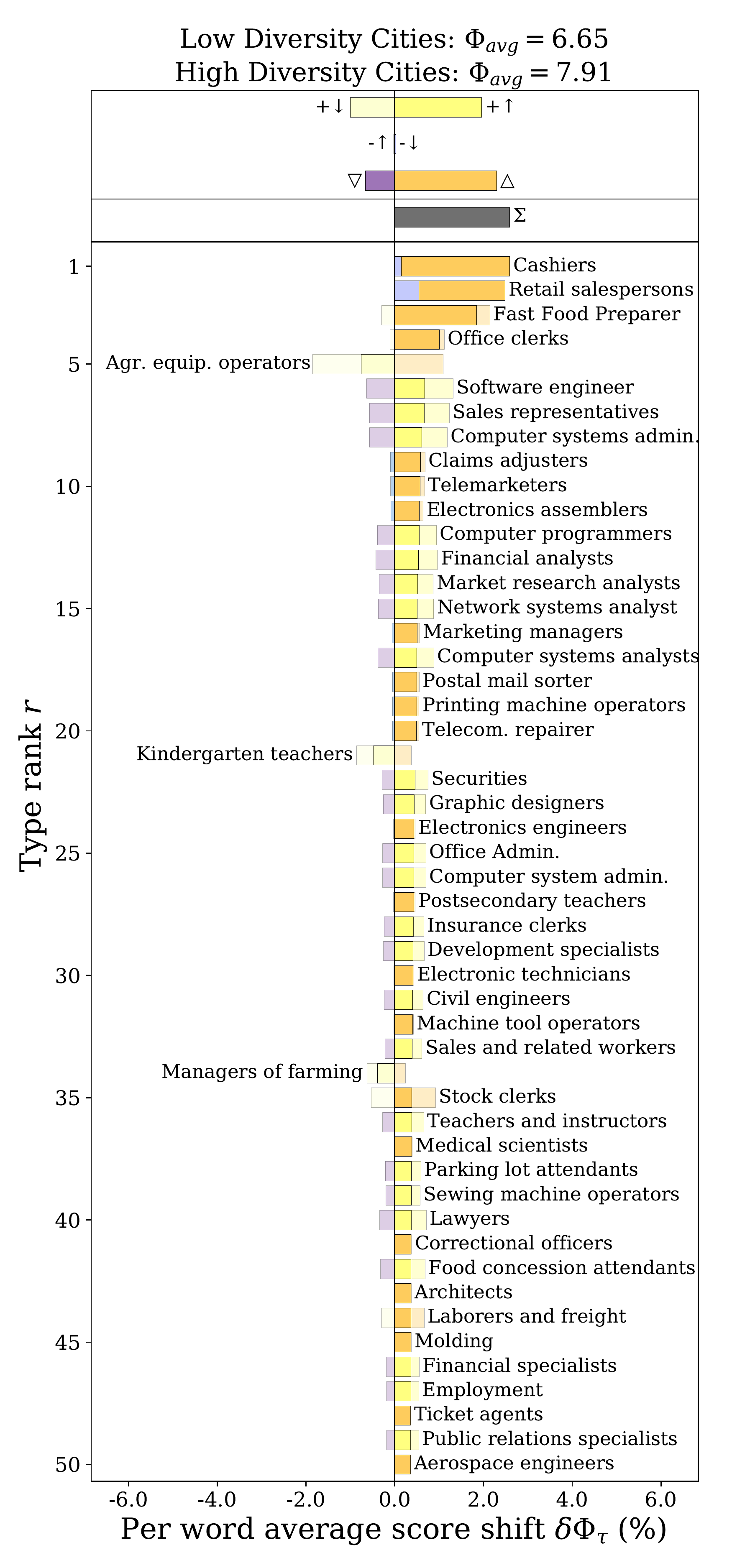}
\caption{
    By treating the employment distribution in each U.S. city as a ``text'' and each occupation as a ``word,'' we employ a generalized word shift graph to compare the differences in employment between the 15 cities with the most diverse employment distributions and the 15 cities with the least diverse employment distributions, as measured by the Shannon entropy.
    We use the average employment diversity across cities' workforces as the reference value ($\Phi^{(\text{ref})}$).
    Occupations are relatively ``surprising'' (+) or ``unsurprising'' (-) depending on their surprisal in each class of city.
}
\label{fig:entropy_shift_labor}
\end{figure}

Traditionally, many labor economists consider job markets to be ``deep'' or ``shallow'' depending on the number of unique occupations, which we refer to as \emph{occupation diversity}, $N^{(c)} = | \mathcal{J}^{(c)} |$. We can more fully account for the distribution of worker employment by considering the \emph{employment diversity}, the Shannon entropy \cite{jost2006entropy,hill1973diversity} of the worker employment distribution,
\begin{equation}
    H \bigl( P^{(c)} \bigr)
        =
            - \sum_{j \in \mathcal{J}} p_j^{(c)} \log_2 p_j^{(c)}.
\end{equation}
Using these measures, we examine the relationship between labor diversity and peak city unemployment during the Great Recession, as given by the BLS's Local Area Unemployment Statistics, and present the results in Fig.~\ref{fig:urbanResilience} in the form of regression coefficients. As we would expect, total employment is significantly positively associated with raw unemployment counts. After controlling for the total employment though, we find evidence that the occupation diversity $N^{(c)}$ is not significantly associated with unemployment, while employment diversity $H^{(c)}$ is significantly negatively associated with peak unemployment during the Great Recession. The inset in Fig.~\ref{fig:urbanResilience} visualizes this negative association when controlling for total employment and occupation diversity.

The relationship between lower unemployment and employment diversity $H^{(c)}$, rather than the number of unique jobs $N^{(c)}$, suggests that urban policy may want to focus on growing the employment diversity of a city's workforce to bolster its economic resilience. However, it is not clear from the aggregate employment diversity which occupations are most likely worth the time, money, and effort of designing policy that reshapes the labor market. By using a (word) shift graph, we can quantify the jobs that most distinguished employment differences between the 15 most diverse cities and the 15 least diverse cities in 2007 (see Fig.~\ref{fig:entropy_shift_labor}). We consider a job to be relatively more ``surprising'' or ``unsurprising'' compared to the entropy of employment distributions averaged across all U.S. cities, i.e. $\langle H^{(c)}\rangle$. As shown by the shift graph, the differences in employment diversity come from two main sources: less common occupations ($+\uparrow$) that are relatively abundant in high diversity cities, and jobs that are common in low diversity cities but less so in high diversity cities $(\uptriangle)$. There are some deviations from these trends; for example, kindergarten teachers and agricultural workers have high surprisal but were relatively more abundant in cities with low employment diversity ($+\downarrow$).

It is beyond the scope of this brief case study to suggest causal policy interventions based on this one shift graph. However, given a more comprehensive study examining labor diversity, unemployment, and economic resilience, it is clear how a shift graph could provide targeted and actionable insights that are not otherwise possible through aggregate measures. Generalized shift graphs are an indispensable tool for enumerating the factors that affect the difference between two weighted averages. These details allow us to more deeply draw on our domain knowledge and build a more comprehensive understanding of the social scientific phenomena under study.

\section{Discussion}

We have introduced generalized word shifts, a framework for taking pairwise comparisons between texts and understanding their differences at the word level. In this framework, comparison measures are decomposed into their individual word contributions so that the words can be ranked and categorized according to how they contribute. The word shift form that we have presented generalizes a previous iteration \cite{dodds2011temporal,reagan2017sentiment}, which was limited to single dictionary-based weighted averages. Our generalization naturally incorporates multi-dictionary scoring, the Shannon entropy, generalized entropies, the Kullback-Leibler divergence, the Jensen-Shannon divergence, and any other measure that can be rewritten as a weighted average or difference in weighted averages. All of these generalized word shifts can be summarily visualized as horizontal stacked bar charts, and we have detailed how to effectively interpret the various interacting components of generalized word shift graphs. To help facilitate their use in computational text analyses, we have implemented generalized word shift graphs in an accessible open source Python package, available at \url{https://github.com/ryanjgallagher/shifterator}.

Generalized word shift graphs are an interpretative tool that allows researchers to fully harness textual measures, both for their audiences and for themselves. While researchers are often limited to arguing in terms of aggregate weighted averages, generalized word shift graphs provide a principled way of decomposing them into word level characterizations that highlight the most salient differences between two texts. In the best case scenario, when the micro word dynamics exposed by a word shift graph align with a macro research story, visualizing word shifts helps audiences better understand and trust what is being measured. However, generalized word shift graphs are not just visual embellishments to persuade audiences. They are also a robustness check that allow us to convince \emph{ourselves} that we have constructed scientifically sound stories. During the research process, generalized word shift graphs can alert us to data peculiarities, counterintuitive phenomena, and measurement errors. Using generalized word shift graphs as a diagnostic tool gives us the opportunity to catch these oddities, account for them, and better understand our text data. Generalized word shift graphs are immediately applicable to a wide range of computational text analyses, and making them a regular part of the text-as-data workflow promises to enrich the work of many computational social scientists, digital humanists, and other practitioners.

Of course, not every text comparison measure can be formulated in terms of weighted averages. For example, many forms of the commonly used term frequency-inverse document frequency cannot be disentangled into a weighted average. Any non-parametric measure that works with ranks rather than frequencies \cite{dodds2020allotaxonometry} cannot, by definition, be written as a weighted average. However, while some additive measures like these cannot be retrofitted into the generalized word shift framework that we have outlined here, we still strongly encourage researchers to always visualize the word contributions that differentiate texts, even if just for themselves during exploratory analyses. Linear, additive text comparison measures are inherently intepretable, and we should always make sure to leverage that interpretability to question, improve, and defend the data stories that we discover.

Generalized word shift graphs directly confront the complexity that is inherent in working with text as data. Used together with other methods, tools, and visualization techniques that open up otherwise opaque black-box methods, word shift graphs can help us better triangulate interesting and meaningful social-scientific phenomenon among the vast and ever expanding landscapes of language, stories, and culture encoded in textual data.

\section{Materials and Methods}

\subsection{Generalized Entropies}

For a relative word frequency distribution $P^{(i)}$, we can calculate its generalized, or Tsallis, entropy of order $\alpha$ \cite{havrda1967quantification,altmann2017generalized},
\begin{align}
    H_\alpha
        &=
            \frac{1}{1 - \alpha}
            \left(
                \sum_{\tau \in \mathcal{T}}
                    p_\tau^\alpha
                -
                1
            \right) \nonumber
        \\
        &=
            \frac{1}{\alpha-1}
            -
            \sum_{\tau \in \mathcal{T}}
                p_\tau^{(i)} 
                \left[ 
                    \frac{
                        p_\tau ^{(\alpha-1)}
                    }{
                        \alpha-1
                    }
                \right],
        \nonumber
\end{align}
where the latter form is more recognizable as a weighted average. The parameter $\alpha$ controls how much weight is given to common and uncommon words. When $\alpha > 1$, more weight is given to frequent words. When $\alpha < 1$, more weight is given to rare words. When $\alpha = 1$, we retrieve the Shannon entropy $H_1 = - \sum_\tau p_\tau^{(i)} \log p_\tau^{(i)}$, which marks the information-theoretic boundary between giving preference to frequently or infrequently occurring words \cite{jost2006entropy,hill1973diversity}. Like the other measures, we can identify a word's contribution by considering the components of $H_\alpha^{(2)} - H_\alpha^{(1)}$,
\begin{equation}
\delta \Phi_\tau
    = 
        - p_\tau^{(2)} 
        \left[ 
            \frac{
                \left(p_\tau^{(2)}\right)^{\alpha-1}
            }{
                \alpha-1
            } 
        \right]
        + p_\tau^{(1)} 
        \left[ 
            \frac{
                \left(p_\tau^{(1)}\right)^{\alpha-1}
            }{
                \alpha-1
            } 
        \right],
\nonumber
\end{equation}
where the quantity $1 / (\alpha -1)$ cancels out in the difference.

The Jensen-Shannon divergence (JSD) can also be extended through generalized entropies \cite{altmann2017generalized}. Recall, for the JSD we form a mixture distribution $M = \pi_1 P^{(1)} + \pi_2 P^{(2)}$, where $\pi_1$ and $\pi_2$ are tunable weights. Rather than writing the JSD as an average KLD relative to the mixture distribution, we can equivalently formulate the JSD in terms of the generalized entropy,
\begin{equation}
    D_\alpha^{(\text{JS})}
        =
            H_\alpha (M) - \pi_1 H_\alpha(P^{(1)}) - \pi_2 H_\alpha(P^{(2)}).
    \nonumber
\end{equation}
With some rearranging, we can write this as a single sum across words and identify their word shift form,
\begin{align}
    \delta \Phi_\tau
        =
            &p_\tau^{(2)}  \pi_2 
            \left[ 
                \frac{
                    \left(p_\tau^{(2)} \right)^{\alpha-1} - m_\tau^{\alpha-1}
                }{
                    \alpha-1
                } 
            \right]
\nonumber 
            \\
             &- p_\tau^{(1)} \pi_1 
             \left[
                \frac{
                    m_\tau^{\alpha-1} - \left(p_\tau^{(1)}\right)^{\alpha-1}
                }{
                    \alpha-1
                }
            \right].
\nonumber
\end{align}
Like the generalized entropy, we recover the familiar JSD when $\alpha = 1$. The heavy tail nature of word distributions can make the JSD sensitive to different word frequencies, particularly when we are working with a sample of texts from a larger corpus (which is very often the case) \cite{altmann2017generalized}. To obtain more reliable estimates of the JSD for those situations, it is advisable to tune the parameter $\alpha$ acccordingly (see ref.~\cite{altmann2017generalized} for details).

\subsection{Derivation of Generalized Word Shifts}

Recall, for a word $\tau$ in text $T^{(i)}$, we denote its relative frequency as $p_\tau^{(i)}$ and its (possibly text dependent) score as $\phi_\tau^{(i)}$. The average score across the entire text $T^{(i)}$ is notated as $\Phi^{(i)}$, and the difference in weighted averages is
\begin{equation}
    \delta \Phi
        =
            \Phi^{(2)} - \Phi^{(1)}
        =
            \sum_{\tau \in \mathcal{T}}
                \phi_\tau^{(2)} p_\tau^{(2)} - \phi_\tau^{(1)} p_\tau^{(1)}.
    \label{eq:derivation-avg-diff}
\end{equation}
We first introduce $\phiref$. Note, $\sum_\tau p_\tau^{(i)} = 1$, and so we may write
\begin{equation}
    \sum_\tau
        \phiref
        \bigl(
            p_\tau^{(2)} - p_\tau^{(1)}
        \bigr)
    =
        \phiref (1 - 1)
    = 
        0.
    \nonumber
\end{equation}
Because the entire above quantity is simply zero, we can subtract it from Eq.~\ref{eq:derivation-avg-diff} to get
\begin{align}
    \delta \Phi
        &=
            \sum_\tau
                \phi_\tau^{(2)} p_\tau^{(2)} - \phi_\tau^{(1)} p_\tau^{(1)}
            - \sum_\tau
                \phiref
                \bigl(
                    p_\tau^{(2)} - p_\tau^{(1)}
                \bigr) \nonumber \\
        &=
            \sum_\tau 
                p_\tau^{(2)} 
                \bigl( 
                    \phi_\tau^{(2)} - \phiref 
                \bigr)
                -
                p_\tau^{(1)} 
                \bigl(
                    \phi_\tau^{(1)} - \phiref 
                \bigr).
    \label{eq:derivation-phiref-form}
\end{align}
Now, we again use the mathematical sleight of hand of adding zero to rewrite the scores $\phi_\tau^{(i)}$ as
\begin{equation}
    \phi_\tau^{(1)}
        =
            \frac{1}{2}
            \biggl(
                \phi_\tau^{(2)} + \phi_\tau^{(1)}
            \biggr)
            -
            \frac{1}{2}
            \biggl(
                \phi_\tau^{(2)}
                -
                \phi_\tau^{(1)}
            \biggr),
    \nonumber
\end{equation}
and
\begin{equation}
    \phi_\tau^{(2)}
        =
            \frac{1}{2}
            \biggl(
                \phi_\tau^{(2)} + \phi_\tau^{(1)}
            \biggr)
            +
            \frac{1}{2}
            \biggl(
                \phi_\tau^{(2)} - \phi_\tau^{(1)}
            \biggr).
    \nonumber
\end{equation}
Substituting into Eq.~\ref{eq:derivation-phiref-form} and working through some algebra, we have the generalized word shift form,
\begin{align}
    \delta \Phi
     =
        \sum_{\tau}
           & \biggl(
                p_\tau^{(2)} - p_\tau^{(1)} 
            \biggr)
            \biggl[
                \frac{1}{2} 
                \bigl(
                    \phi_\tau^{(2)} + \phi_\tau^{(1)} 
                \bigr)
                - \phiref
            \biggr] \nonumber \\
            &+
            \frac{1}{2}
            \biggl(
                p_\tau^{(2)} + p_\tau^{(1)}
            \biggr)
            \biggl(
                \phi_\tau^{(2)} - \phi_\tau^{(1)}
            \bigg).
    \nonumber
\end{align}
The basic word shift \cite{dodds2011temporal} is a special case of the general form when the scores are text independent $\phi_\tau^{(1)} = \phi_\tau^{(2)} = \phi_\tau$,
\begin{equation}
    \delta \Phi
    =
        \sum_{\tau}
            \biggl(
                p_\tau^{(2)} - p_\tau^{(1)} 
            \biggr)
            \biggl(
                \phi_\tau - \phiref
            \biggr).
    \nonumber
\end{equation}

\subsection{Handling Missing Types and Scores}

At times, we may have a word present in one text only, and so either $p_\tau^{(1)} = 0$ and $p_\tau^{(2)} > 0$, or $p_\tau^{(2)} = 0$ and $p_\tau^{(1)} > 0$. If the scores $\phi_\tau^{(i)}$ are functions of the relative frequencies, then this can be problematic at times. Consideration needs to be given to the particular measure at hand to decide how to deal with the missing types. For some of the measures, like the generalized entropy, setting $ \phi_\tau^{(i)} = 0$ does not cause any mathematical troubles. For the Shannon entropy, it may seem at first that we have cause for concern because $\phi_\tau^{(i)} = -\log p_\tau^{(i)}$, which is undefined if $p_\tau^{(i)} = 0$. However, in the Shannon entropy the surprisal is always multiplied by $p_\tau^{(i)}$, and so by the magic of limits and differential calculus, we can safely write
\begin{equation}
    -p_\tau^{(i)} \log p_\tau^{(i)} = 0,
\end{equation}
when $p_\tau^{(i)} = 0$. Practically, for programmatic implementation, it is enough to set $\phi_\tau^{(i)} = 0$. Similarly for the JSD, the form of $\phi_\tau^{(1)} = \log m_\tau / p_\tau^{(1)}$ may appear to be undefined when $p_\tau^{(1)} = 0$. If we step back to the overall word contribution $\delta \Phi_\tau$ shown in Eq.~\ref{eq:jsd-contribution} though, then by the same limiting argument as the Shannon entropy, we can safely simply set $\phi_\tau^{(1)} = 0$. The same cannot be done for the KLD though, unfortunately. One part of the word contribution is the quantity $-p_\tau^{(2)} \log p_\tau^{(1)}$. Because the relative frequencies are different in this case (unlike the Shannon entropy and JSD), no amount of applying L'H\^opital's rule will give us a finite limit as $p_\tau^{(1)}$ approaches zero. All we can say is that the KLD is undefined when $p_\tau^{(1)} = 0$ and $p_\tau^{(2)} > 0$.

For dictionary scores, we may have both $p_\tau^{(1)} > 0$ and $p_\tau^{(2)} > 0$, but, without loss of generality, only $\phi_\tau^{(1)}$ is defined. This can happen in domain-adapted sentiment dictionaries. For example, if we were to build sentiment dictionaries for every year since the early 1900s, we would not be able to assign sentiment to the hashtag ``\#worldwar3'' for texts from 1910 because neither social media nor World War I existed at the time. In a less extreme case, ``trump'' is certainly used enough in 2020 to be included in any contemporary frequency-based sentiment dictionary, but it may not have been used enough for inclusion in a dictionary for the 1940s, even though the word certainly existed. There is ambiguity in how to handle these cases. We default to setting the missing score $\phi_\tau^{(2)}$ to the value of the defined score $\phi_\tau^{(1)}$. In practice, this nullifies the score difference component, and places the contribution's emphasis on the basic word shift component. This may be reasonable in some cases and less so in others, particularly if we expect a shift in a word's sentiment to be well-defined and noticeable between two texts (as may be the case with ``trump'' between 1940 and 2020). If that is the case, it may be necessary to further expand the sentiment dictionaries, or exert domain expertise to make some other defensible decision.

\subsection{Case Studies}

\subsubsection{Presidential Speeches}

We collected presidential speeches online from the University of Virginia's Miller Center (\url{https://millercenter.org/the-presidency/presidential-speeches}). The text of each speech is clearly organized by speaker and we parsed them to separate presidents from other entities (such as audiences or moderators). Unigrams were lowercased and the average sentiment was calculated over a president's entire set of speeches as a single text (not the average of average sentiments of each individual speech). Our dataset includes 71 speeches from Lyndon B. Johnson, consisting of 256,133 word tokens across 10,094 word types, and 39 speeches from George W. Bush, consisting of 107,913 tokens across 7,804 types. For the labMT sentiment dictionary \cite{dodds2011temporal}, we use a reference value of $\phiref = 5$, which is the center of the dictionary's 1 to 9 sentiment scale. We also apply a stop window which excludes any labMT word whose sentiment falls between the scores 4 and 6. For the SocialSent historical lexicons \cite{hamilton2016inducing}, we use a reference value of $\phiref = 0$, as all dictionaries were scaled to have a mean of zero when they were constructed.

\begin{table*}[!htbp]
    \centering
\begin{tabular}{c|ccccc}
\hline
Variable& Model 1& Model 2& Model 3& Model 4& Model 5\\ \hline
$\log_{10}$ Total Employment ($T^{(c)}$)& $0.950^{***}$& & & $1.214^{***}$& $1.296^{***}$\\
Employment Diversity ($H^{(c)}$)& & $0.802^{***}$& & $-0.220^{***}$& $-0.203^{***}$\\
Occupation Diversity ($N^{(c)}$)& & & $0.925^{***}$& $-0.068$& $-0.140$\\
$T^{(c)}\times H^{(c)}$& & & & & $0.114$\\
$N^{(c)}\times H^{(c)}$& & & & & $-0.040$\\
$T^{(c)}\times N^{(c)}$& & & & & $-0.072^{*}$\\\hline
\rule{0pt}{3ex}$R^2$& 0.903& 0.643& 0.856& 0.914& 0.915\\ \hline
\rule{0pt}{3ex}adj. $R^2$& 0.903& 0.642& 0.856& 0.913& 0.914\\ \hline
\multicolumn{6}{c}{$p_{val}<0.1^*$, $p_{val}<0.01^{**}$, $p_{val}<0.001^{***}$} \\ \hline
\end{tabular}
\caption{
    Regressing urban labor statistics against $\log_{10}$ the maximimum unemployment in each U.S. city during the Great Recession.
    All variables are centered and standardized prior to analysis.
    Regression coefficient estimates from Model 4 are presented in Fig.~\ref{fig:urbanResilience}.
}
\label{table:cityRegression}
\end{table*}

\subsubsection{Moby Dick}

The raw text of Moby Dick by Herman Melville is freely available on Project Gutenberg at \url{http://www.gutenberg.org/files/2701/2701.txt}.
We process the raw text by removing the head matter and manually ending the text at the `ETYMOLOGY' section.
For the figures in this paper,
we use a manually trimmed version of the raw text,
with chapter headings removed (in contrast the larger emotional arc corpus \cite{reagan2016emotional}, which relied on automated header and footer removal).
We remove spaces and punctuation, and lowercased all tokens.
There are 213,984 total tokens in Moby Dick across 16,858 word types, resulting in 106,992 tokens in each the first and second halves with 11,930 and 11,646 word types, respectively.
For sentiment scores, we make the same choices as we did for the presidential speeches: we use the labMT sentiment dictionary \cite{dodds2011temporal}, apply a stop window which excludes any labMT word whose sentiment falls between the scores 4 and 6, and use a reference value of 5.
A reproducible analysis is available at \url{https://github.com/andyreagan/shifterator-case-study-moby-dick} (as mentioned above, the results herein rely on the `raw' versions in the codebase).

\subsubsection{US Urban Parks}
We collected tweets from Twitter's Decahose (10\%) feed, stored in the Computational Story Lab's database at the University of Vermont. We restricted our sample to English language tweets with GPS coordinates posted from January 1st, 2012 to April 27th, 2015 (a period in which geolocation was widely used). Using boundaries from the US Census, we subsampled tweets within each of the 25 largest cities in the US by population. Within these cities, we found 297,494 posted within urban park boundaries using the Trust for Public Land's Park Serve database at \url{https://parkserve.tpl.org/}. To compare sentiment between in-park and out-of-park tweets, we paired each in-park tweet with the closest-in-time out-of-park tweet from another user within the same city (see ref.~\cite{2020arXiv200610658S} for details). Across the park tweets, there were 3,920,722 tokens across 451,627 word types. Across the out-of-park control tweets there were 3,861,357 tokens across 410,397 word types. For sentiment scores, we make the same choices as we did for the presidential speeches and Moby Dick: we use the labMT sentiment dictionary \cite{dodds2011temporal}, apply a stop window which excludes any labMT word whose sentiment falls between the scores 4 and 6, and use a reference value of 5.

\subsubsection{Information Content of 280 Character Tweets}

We collected English-language tweets from Twitter's decahose (10\%) feed, stored in the Computational Story Lab's database at the University of Vermont.
Language detection came from the `en' language label on each tweet provided by Twitter's API. 
This comprised 577,985,080 tweets over the 60-day period studied:
274,888,052 from the 30 days before 7 November 2017, and 303,097,028 from the 30 days afterwards.
We restricted to considering changes in a consistent vocabulary of all 10,222 word types contained in the LabMT dictionary (i.e., without removal of any stop words) before and after the change.
This resulted in a collection of 2,526,152,975 word tokens from the period before the change, and 2,555,503,284 from the period after the change.
We use the average entropy of 140 character tweets as the reference value for the generalized word shift.

\subsubsection{Regression Analysis of Urban Labor Diversity and the Great Recession}

Employment data for U.S. cities in 2007 comes from Occupational Employment Statistics (OES) data provided by the U.S. Bureau of Labor Statistics (BLS). 
Employment is reported using the Standard Occupation Classification (SOC) system that unifies occupational data across the U.S. Department of Labor.
The SOC is a hierarchical classification system, and we use the most detailed (i.e., 6-digit) occupation codes in our analysis.
However, occupation titles in Fig.~\ref{fig:entropy_shift} are simplified to conserve space; for example, the occupation category ``Correctional Officers and Jailers'' (occupation code: 33-3012) is simplified to ``Correctional officers.''
For comparing high and low diversity cities in Fig.~\ref{fig:entropy_shift}, we first rank U.S. cities based on the Shannon entropy of their employment distributions in 2007 (i.e., $H^{(c)}$) and consider the 15 most diverse cities to the 15 least diverse cities.
For eeach one of these collections of 15 cities, we produce an aggregated employment distribution by taking the average employment share for each occupation across the cities in the collection of cities.

We analyze unemployment in U.S. cities during the Great Recession using Local Area Unemployment Statistics (LAUS) provided from the U.S. BLS.
This data includes monthly statistics for each U.S. city.
Since economic disruptions begin in different cities at different times and urban economies recover at different rates, we consider the month in the period between January 2008 and December 2012 with the most unemployment in a given city.

Table~\ref{table:cityRegression} displays a more complete analysis of urban labor statistics and unemployment during the Great Recession.
All variables are centered and standardized prior to analysis so that each variable is unit-less; this makes it easier to compare the relative importance of each independent variables in predicting the dependent variable.
It is very important to first control for the size of each city's labor force (i.e., $T^{(c)}$) before considering the effects of labor diversity on economic resilience.
This is because cities with larger labor forces have greater potential for absolute unemployment.
Models 1, 2, and 3 show the Pearson correlations between each individual independent variable and the dependent variable.
Model 3 combines all independent variables and reveals that both $T^{(c)}$ and $H^{(c)}$ are significant predictors of maximum unemployment during the Great Recession, but occupation diversity (i.e., $N^{(c)}$) is not.
Adding the measures for labor diversity in addition to labor force size yields an improvement in the overall predictive performance of the regression model from 90.3\% variance explained to 91.4\% thus accounting for an additional 14\% of the unexplained variance when using labor force size alone.
Finally, Model 5 includes the interaction terms between independent variables and again demonstrates the added predictive value of $H^{(c)}$ in addition to $T^{(c)}$.
Interestingly, we also find large cities with large occupation diversity experienced lower unemployment during the Great Recession.


\section*{Acknowledgements}

We thank Brandon Stewart and the other the attendees of the Ninth Annual Conference on New Directions in Analyzing Text as Data for their useful feedback on the presentation of word shift graphs. CMD and PSD are grateful for financial support from the Massachusetts Mutual Life Insurance Company and Google Open Source under the Open-Source Complex Ecosystems And Networks (OCEAN) project. Computations were performed on the Vermont Advanced Computing Core supported in part by NSF award No. OAC-1827314.


\bibliography{bibl}
\bibliographystyle{unstrabbrv}

\end{document}